\documentclass[runningheads]{llncs}
\usepackage{graphicx}

\usepackage{tikz}
\usepackage{comment}
\usepackage{amsmath,amssymb} 
\usepackage{color}

\usepackage{booktabs}
\usepackage{graphics}
\usepackage{appendix}

\usepackage[accsupp]{axessibility}

\usepackage[pagebackref,breaklinks,colorlinks]{hyperref}

\begin{document}
\pagestyle{headings}
\mainmatter

\title{TISE: Bag of Metrics for Text-to-Image Synthesis Evaluation}

\titlerunning{TISE: Text-to-Image Synthesis Evaluation}
\author{Tan M. Dinh\thanks{Corresponding author}\index{Dinh, Tan M.} \and Rang Nguyen \and Binh-Son Hua\index{Hua, Binh-Son}}
\authorrunning{T. Dinh et al.}
\institute{VinAI Research, Hanoi, Vietnam}

\maketitle

\begin{figure}[t]
\centering
\includegraphics[width=0.8\linewidth]{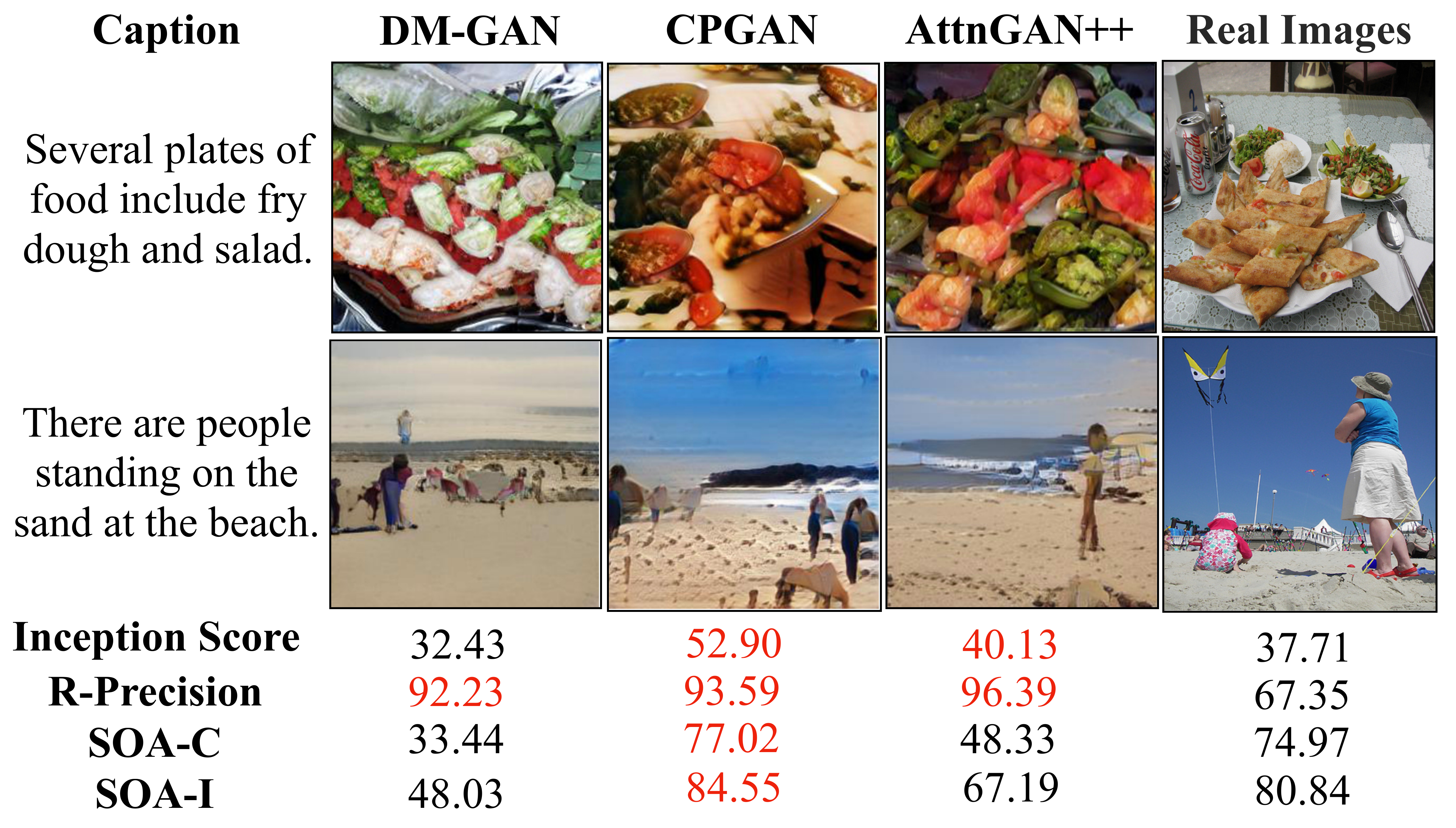}
\caption{Evaluating the text-to-image models is a challenging task. Many existing metrics are inconsistent especially for the case when an input sentence involves multiple objects. Values in \textcolor{red}{red} denote inconsistent evaluations, where the quantitative results are even higher than that of real photos, despite the fact that such generated images are not perceptually real.}
\label{fig:teaser}
\end{figure} 

\begin{abstract}
In this paper, we conduct a study on the state-of-the-art methods for text-to-image synthesis and propose a framework to evaluate these methods. We consider syntheses where an image contains a single or multiple objects. Our study outlines several issues in the current evaluation pipeline: (i) for image quality assessment, a commonly used metric, e.g., Inception Score (IS), is often either miscalibrated for the single-object case or misused for the multi-object case; (ii) for text relevance and object accuracy assessment, there is an overfitting phenomenon in the existing R-precision (RP) and Semantic Object Accuracy (SOA) metrics, respectively; (iii) for multi-object case, many vital factors for evaluation, e.g., object fidelity, positional alignment, counting alignment, are largely dismissed; (iv) the ranking of the methods based on current metrics is highly inconsistent with real images. To overcome these issues, we propose a combined set of existing and new metrics to systematically evaluate the methods. For existing metrics, we offer an improved version of IS named IS* by using temperature scaling to calibrate the confidence of the classifier used by IS; we also propose a solution to mitigate the overfitting issues of RP and SOA. For new metrics, we develop counting alignment, positional alignment, object-centric IS, and object-centric FID metrics for evaluating the multi-object case. We show that benchmarking with our bag of metrics results in a highly consistent ranking among existing methods that is well-aligned with human evaluation. As a by-product, we create AttnGAN++, a simple but strong baseline for the benchmark by stabilizing the training of AttnGAN using spectral normalization. We also release our toolbox, so-called TISE, for advocating fair and consistent evaluation of text-to-image models.
\end{abstract}

\section{Introduction}
The unprecedented growth of deep learning has sparked significant interest in tackling the vital vision-language task of text-to-image synthesis in recent years, with potential applications from computer-aided design, image editing with text-guided to image retrieval. This is a challenging task because of the wide semantic gap between two domains and the high many-to-many mapping (e.g., one text caption can correspond to many image counterparts and vice versa). Many aspects of image synthesis, such as image fidelity, object relations, object counting have to be considered for generating complex scenes from a sentence. 
	
In the past few years, key techniques for text-to-image synthesis are largely based on the evolution of generative adversarial networks (GANs)~\cite{goodfellow2014generative}. Tremendous achievements has been obtained in many domains, e.g. from unconditional image generation~\cite{karras2019style,karras2020analyzing} to latent space mapping and manipulation~\cite{shen2020interpreting,shen2020interfacegan}.   Most of text-to-image synthesis approaches~\cite{xu2018attngan,li2019object,tan2019semantics,yin2019semantics,zhu2019dm,liang2019cpgan} are built upon GANs and jointly consider text and image features in the synthesis.

Despite excellent results have been achieved on particular datasets~\cite{Nilsback08,welinder2010caltech,lin2014microsoft}, the current evaluation pipeline is far from ideal. For single object case, image quality and text-image alignment are primary factors considered in a typical evaluation process. Some commonly evaluation metrics are Inception Score (IS)~\cite{salimans2016improved} and the Fr\'echet Inception Distance (FID)~\cite{heusel2017gans} for image fidelity and R-precision (RP)~\cite{xu2018attngan} for text-image alignment, which works well for most single-object cases. However, in complex scenes with multiple objects, adopting these metrics are not enough and causes some inconsistency issues. As can be seen in Figure~\ref{fig:teaser}, the ranking of GAN models based on the current metrics is not strongly correlated to their generated image qualities. The numbers reported from several GANs are even better than the one of corresponding real images, while it is clearly seen that the quality of generated images are still far from being real. Additionally, the existing evaluation system lacks the metrics for assessing other aspects like object fidelity, positional alignment, and counting alignment, among others. These aspects are critical in evaluating the performance of text-to-image models in the multi-object case. Furthermore, the absence of a unified evaluation toolbox has resulted in inconsistent outcomes reported by different research works. These issues are also highlighted in the recent comprehensive survey \cite{frolov2021adversarial}, which raises a demand to devise a unified bag of metrics for text-to-image evaluation.

In this paper, we develop a systematic method for evaluating text-to-image synthesis approaches to tackle the challenges mentioned above. Our contributions are summarized as follows:

\begin{enumerate}

\item For existing metrics, we create IS* as an improved version of IS metric for image quality assessment, which alleviates the low confidence phenomenon due to miscalibrations in the pre-trained classifier used for IS. We also develop the robust versions for text relevance and object accuracy assessment (RP and SOA~\cite{hinz2019semantic}) to mitigate their overfitting issues in multi-object case. 

\item For new metrics, we develop O-IS and O-FID for object fidelity, PA for positional alignment, and CA for counting alignment to evaluate these lacking aspects in multi-object text-to-image synthesis. 
	
\item Based on these metrics, we conduct a comprehensive, fair and consistent evaluation of the current state-of-the-art methods for both single- and multi-object text-to-image models.

\item Finally, we propose AttnGAN++, a simple but strong baseline that works well for both single- and multi-object scenarios. Our AttnGAN++ has competitive performance to current state-of-the-art text-to-image models.

\end{enumerate}

On top of these contributions, we develop a \emph{Python} assessment toolbox called \textbf{TISE} (\underline{\textbf{T}}ext-to-\underline{\textbf{I}}mage \underline{\textbf{S}}ynthesis \underline{\textbf{E}}valuation) implementing our bag of metrics in a unified way to facilitate, advocate fair comparisons and reproducible results for future text-to-image synthesis research. 
\footnote{TISE toolbox is available at \url{https://github.com/VinAIResearch/tise-toolbox}.} 

\section{Background}
\label{sec:background}
\noindent \textbf{\emph{Text-to-Image Synthesis}} is a vision-language task substantially benefit from the unprecedented evolutions of generative adversarial neural networks and language models. GAN-INT-CLS \cite{reed2016generative} is the first conditional GAN~\cite{mirza2014conditional} designed for text-to-image generation, but images generated by GAN-INT-CLS only have $64 \times 64$ resolution. StackGAN and its successor StackGAN++ \cite{zhang2017stackgan,zhang2018stackgan++} enhanced the resolution of generated images by using a multi-stage architecture. These works, however, only consider sentence-level features for image synthesis; word-level features are completely dismissed, which causes poor image details. To fix this issue, an attention mechanism can be used to provide word-level features, notably used by AttnGAN \cite{xu2018attngan} and DM-GAN \cite{zhu2019dm}, which significantly improves the generated image quality. Beyond modifying the network architecture, improving semantic consistency between image and caption is also an active research topic to gain better image quality. SD-GAN \cite{yin2019semantics} and SE-GAN \cite{tan2019semantics} guarantee text-image consistency by the Siamese mechanism; \cite{qiao2019mirrorgan} proposes a text-to-image-to-text framework called MirrorGAN inspired by the cycle consistency, while \cite{zhang2021cross,ye2021improving} leverage contrastive learning in their text-to-image models. To improve the performance of model in the multi-object case, InferGAN \cite{hong2018inferring} and Obj-GAN \cite{li2019object} introduce a two-step generation process including layout generation and image generation, while CPGAN \cite{liang2019cpgan} leverages the object memory features in developing the model. Regarding model scaling approach, DALL-E \cite{pmlr-v139-ramesh21a} and CogView~\cite{ding2021cogview} are two large scale text-to-image synthesis models with $12$ and $4$ billion parameters, respectively, synthesizing the image from the caption autoregressively by using a transformer~\cite{vaswani2017attention} and VQ-VAE~\cite{razavi2019generating}. 
	
\noindent\textbf{\emph{Evaluation.}} The rapid advancement of text-to-image generation necessitates the construction of a reliable and systematic evaluation framework to benchmark models and guide future research. However, assessing the quality of generative modeling tasks has proven difficult in the past~\cite{theis2015note}. Because none of the existing measures are perfect, it is usual to report many metrics, each of which assesses a different aspect. The performance assessment is even more challenging in the text-to-image synthesis task due to the multi-modal complexity of text and image, which motivates us to develop a new evaluation toolbox to compare text-to-image approaches fairly and confidently.
	
\section{Single-Object Text-to-Image Synthesis}
\subsection{Existing Metrics} 
Most of existing metrics access the quality of model based on two aspects: image quality and text-image alignment. For assessing the image quality of the model, Inception score (IS) \cite{salimans2016improved} and Fréchet Inception Distance (FID) \cite{heusel2017gans} are two common metrics. These metrics originally come from traditional GAN tasks for evaluating the image quality. For evaluating text-image alignment, R-precision~\cite{xu2018attngan} metric is utilized popularly.

\noindent\textbf{\emph{Inception Score (IS)}}~\cite{salimans2016improved} leverages a pretrained Inception-v3 network \cite{szegedy2016rethinking} for calculating the Kullback-Leibler divergence (KL-divergence) between class-conditional distribution and class-marginal distribution of the generated images. The formula of IS is defined below.
\begin{equation} 
\mathrm{IS} = \exp({\mathop{{}\mathbb{E}}}_{x} D_{KL} (p(y|x) \parallel p(y))),
\end{equation}
where $x$ is the generated image and $y$ is the class label.
The goal of this metric is to determine whether a decent generator can generate samples under two conditions: \emph{(i)} The object in the image should be \emph{distinct} $\rightarrow$ $p(y|x)$ must have low entropy; \emph{(ii)} Generated images should have the \emph{diversity} of object class $\rightarrow$ $p(y)$ must have high entropy. Combining these two considerations, we expect that the KL-divergence between $p(y)$ and $p(y|x)$ should be large. Therefore, higher IS value means better image quality and diversity. 

\noindent\textbf{\emph{Fréchet Inception Distance (FID)}}~\cite{heusel2017gans} calculates the Fréchet distance between two sets of images: generated and actual. To calculate FID, features from each set are firstly extracted by a pre-trained Inception-v3 network \cite{szegedy2016rethinking}. Then, these two feature sets are modeled as two \emph{multivariate Gaussian distributions}. Finally, the Fréchet distance is calculated between two distributions.
\begin{equation}
\mathrm{FID} = || \mu_{r} - \mu_{g} ||^{2} + \mathrm{trace}\left( \Sigma_{r} + \Sigma_{g} - 2(\Sigma_{r} \Sigma_{g})^{\frac{1}{2}} \right), 
\end{equation}
where \(X_{r} \sim \mathcal{N}(\mu_{r}, \Sigma_{r})\) and  \(X_{g} \sim \mathcal{N}(\mu_{g}, \Sigma_{g})\) are the features of real images and generated images extracted by a pretrained Inception-v3 model. Lower FID value means better image quality and diversity. 
	
\noindent\textbf{\emph{R-precision (RP)}}~\cite{xu2018attngan} metric is used popularly to evaluate text-image consistency. The idea of RP is to use synthesized image query again the input caption. In particular, given a ground truth text description and $99$ mismatching captions sampled randomly, an image is generated from ground truth caption. Then this image is used to query again input description among $100$ candidate captions. This retrieval is marked as successful if the matching score of it and ground truth caption is the highest one. The cosine similarity between image encoding vector and caption encoding vector is used as matching score. RP is the ratio of successful retrieval and higher score means better quality.
	
\subsection{Benchmark Results}

\setlength{\tabcolsep}{4pt}
\begin{table}[t]
\begin{center}
\caption{Benchmark results for the single-object text-to-image synthesis models on the CUB dataset. In this benchmark, we only consider the methods, which have been released with officially source code and pre-trained weights by their authors. \textbf{Best} and \underline{runner-up} values are marked in bold and underline.}
\label{tab:cub_evaluation}
\begin{tabular}{l c c c} 
\toprule
Method & IS ($\uparrow$) & FID ($\downarrow$) & RP ($\uparrow$) \\
\midrule
GAN-INT-CLS~\cite{reed2016generative} & 2.73 & 194.41 & 3.83 \\

StackGAN++~\cite{zhang2018stackgan++} & 4.10  & 27.40 & 13.57 \\

AttnGAN~\cite{xu2018attngan} & 4.32 & 24.27 & 65.30 \\ 

AttnGAN + CL~\cite{ye2021improving}  &  4.45  & 17.96  & 60.82 \\

DM-GAN~\cite{zhu2019dm} & 4.68 & 15.52 & \underline{76.25}  \\

DF-GAN~\cite{ming2020DFGAN} &  \underline{4.77} & 16.46  & 42.95 \\

DM-GAN + CL~\cite{ye2021improving}   & \underline{4.77}  &  \textbf{14.57}  &  69.80 \\

\midrule

AttnGAN++ (ours) & \textbf{4.78}  & \underline{15.01} & \textbf{77.31} \\

\bottomrule
\end{tabular}
\end{center}
\end{table}
	
In this section, we conduct an assessment to re-evaluate existing text-to-image models in the single-object case. For simplicity, CUB dataset \cite{welinder2010caltech} is selected for our mini-benchmark and used to generate images with only one object from fine-grained text description. CUB dataset \cite{welinder2010caltech} contains $11,788$ images from $200$ different bird species. We follow the same setup as mentioned in~\cite{zhang2017stackgan} to pre-process and prepare train/test data in zero-shot setting. 
	
We suggest a new baseline approach for this benchmark based on recent breakthroughs in deep learning techniques, in addition to previous efforts. Particularly, we revise the architecture of AttnGAN \cite{xu2018attngan} by adding the spectral normalization layers to the discriminator that helps stabilize the training process. We also hand-tune the hyperparameters of our baseline network, which we denote as AttnGAN++. The detail architecture and network setting of AttnGAN++ are shown in supplementary material. The quantitative results of our benchmark is reported in Table~\ref{tab:cub_evaluation}, which brings the following insights. \\
    
\noindent\textbf{\emph{Insight 1: AttnGAN++ is a strong baseline.}} As can be seen in  Table~\ref{tab:cub_evaluation}, our AttnGAN++ outperforms  the original version (AttnGAN) with a large gap on all metrics for CUB dataset and has the comparable results with existing state-of-the-art works. It is worth noting that most of current state-of-the-art works~\cite{zhu2019dm,ye2021improving,liang2019cpgan,li2019control} are built on AttnGAN. Therefore, this empirical finding would help create a very strong baseline for further improving the successor works. The qualitative results can be found in the supplementary. 
\begin{figure}[t]
\begin{centering}
\noindent\begin{minipage}[t]{1\columnwidth}%
\begin{center}
\begin{minipage}[t]{0.45\columnwidth}%
\begin{center}
\includegraphics[width=0.9\columnwidth,height=0.9\columnwidth]{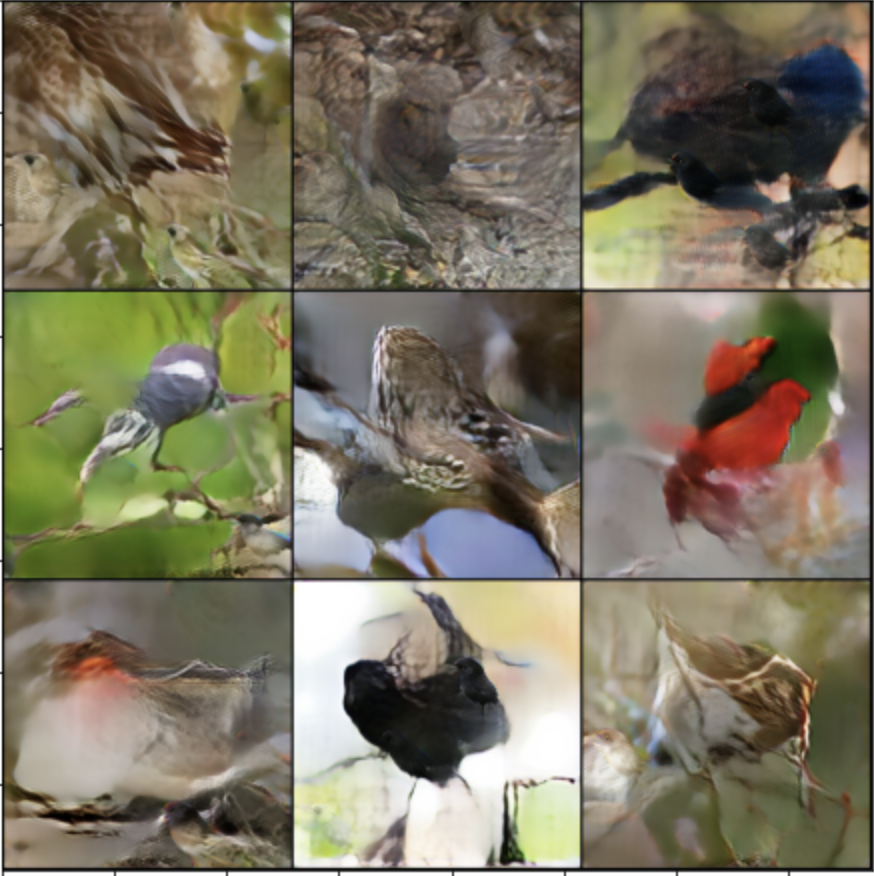}\\
(a) $\mathrm{IS = 5.12}$; $\mathrm{IS^{*} = 13.05}$
\par\end{center}%
\end{minipage}\quad{}%
\begin{minipage}[t]{0.45\columnwidth}%
\begin{center}
\includegraphics[width=0.9\columnwidth,height=0.9\columnwidth]{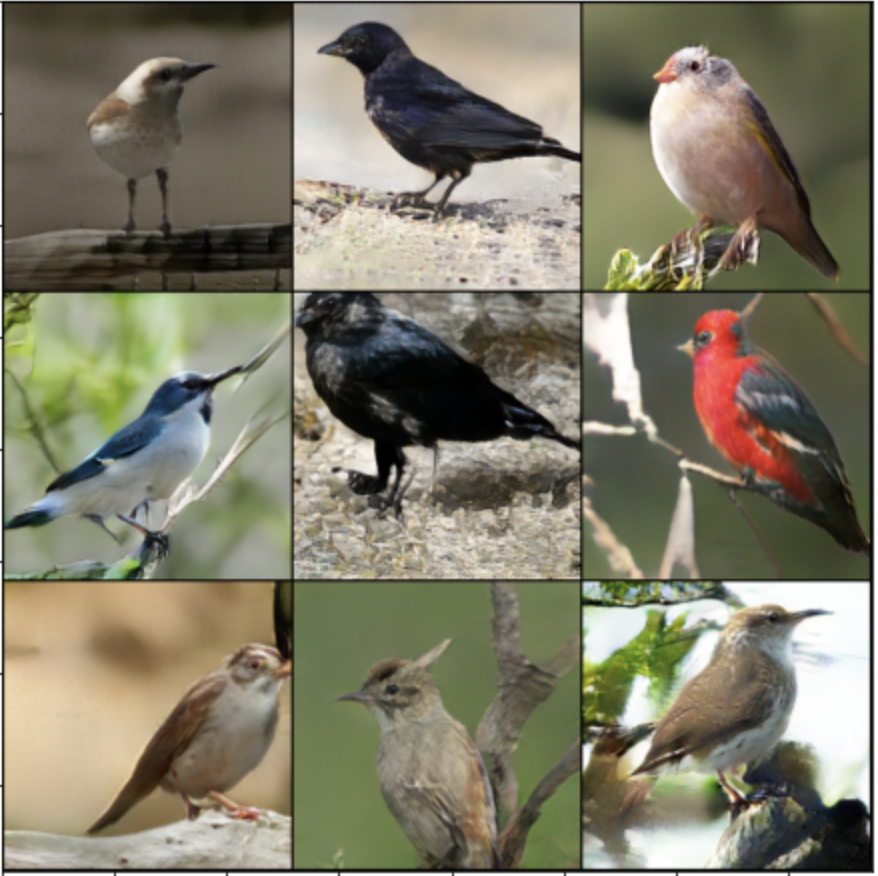}\\
(b) $\mathrm{IS = 4.78}$; $\mathrm{IS^{*} = 15.13}$
\par\end{center}%
\end{minipage}
\par\end{center}%
\end{minipage}
\par\end{centering}
\caption{Evaluating the  single-object text-to-image synthesis models can be inconsistent with the IS score. (a) Generated images from the counter model are unrealistic but the IS score of this model is high; (b) Generated images of our AttnGAN++. As can be seen, our IS* fixes well this inconsistency issue.}
\label{fig:counter}
\end{figure}

\noindent\textbf{\emph{Insight 2: IS scores are inconsistent.}} During the development of AttnGAN++, we discovered that it is feasible to design a generator that produces unrealistic images while yet having a high IS score, which we refer to as the \emph{counter model}. Generated images from this counter model is shown in Figure~\ref{fig:counter}(a). Note that the images from the counter model are randomly sampled and not curated. 
We describe the architecture of this counter model as well as how to reproduce these results in the supplementary. 
    
Motivated by Insight 2, we revisited the definition of IS metric, and discovered that the inconsistency is due to a pitfall when the IS score is computed in the text-to-image synthesis task. From this observation, we proposed an improved version of IS that address such limitation, as follows.

\subsection{Improved Inception score (IS*): Calibrating Image Classifiers} 

\begin{figure}[t]
\begin{centering}
\noindent\begin{minipage}[t]{1\columnwidth}%
\begin{center}
\begin{minipage}[t]{0.45\columnwidth}%
\begin{center}
\includegraphics[width=0.9\columnwidth,height=0.9\columnwidth]{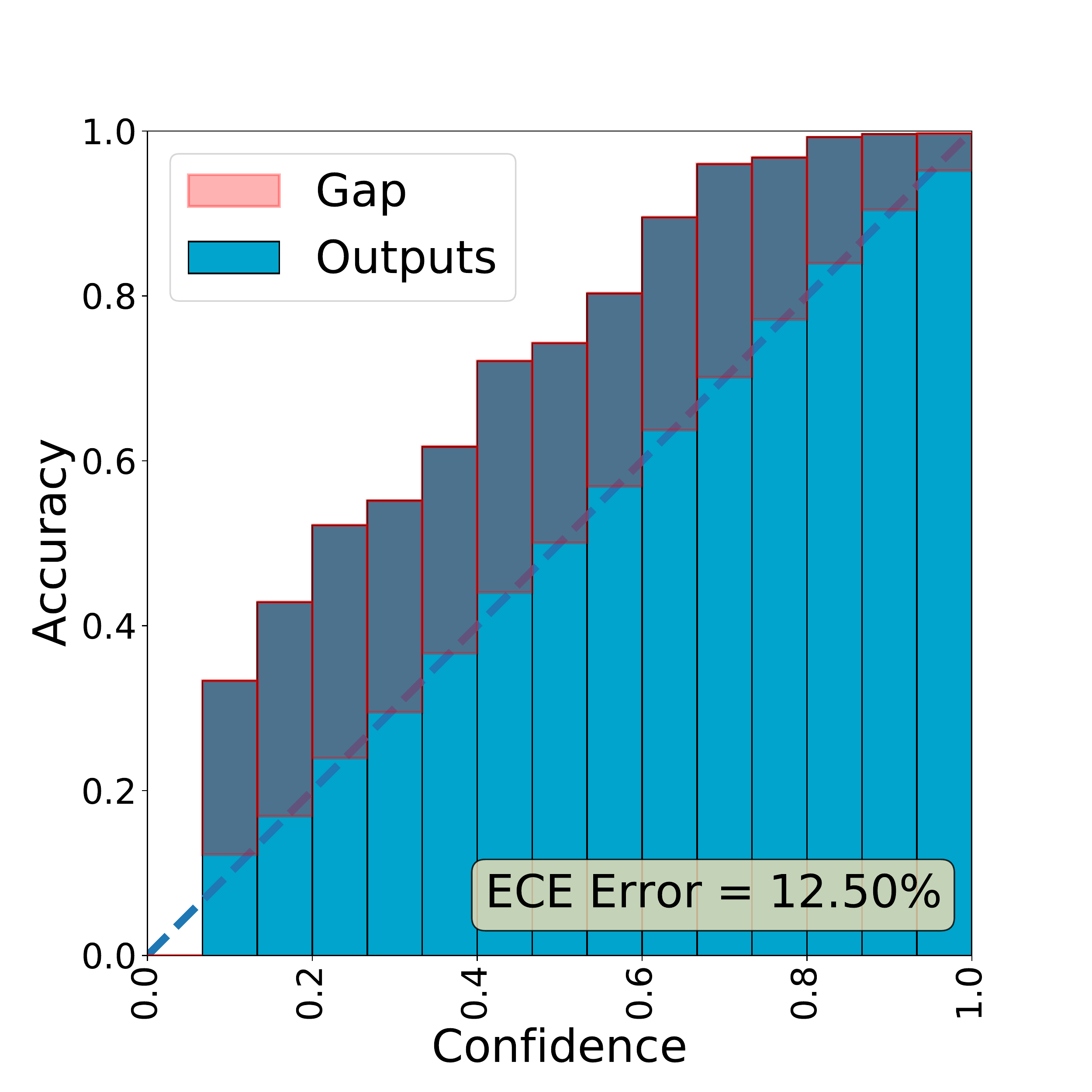}\\
(a) Before calibration
\par\end{center}%
\end{minipage}\quad{}%
\begin{minipage}[t]{0.45\columnwidth}%
\begin{center}
\includegraphics[width=0.9\columnwidth,height=0.9\columnwidth]{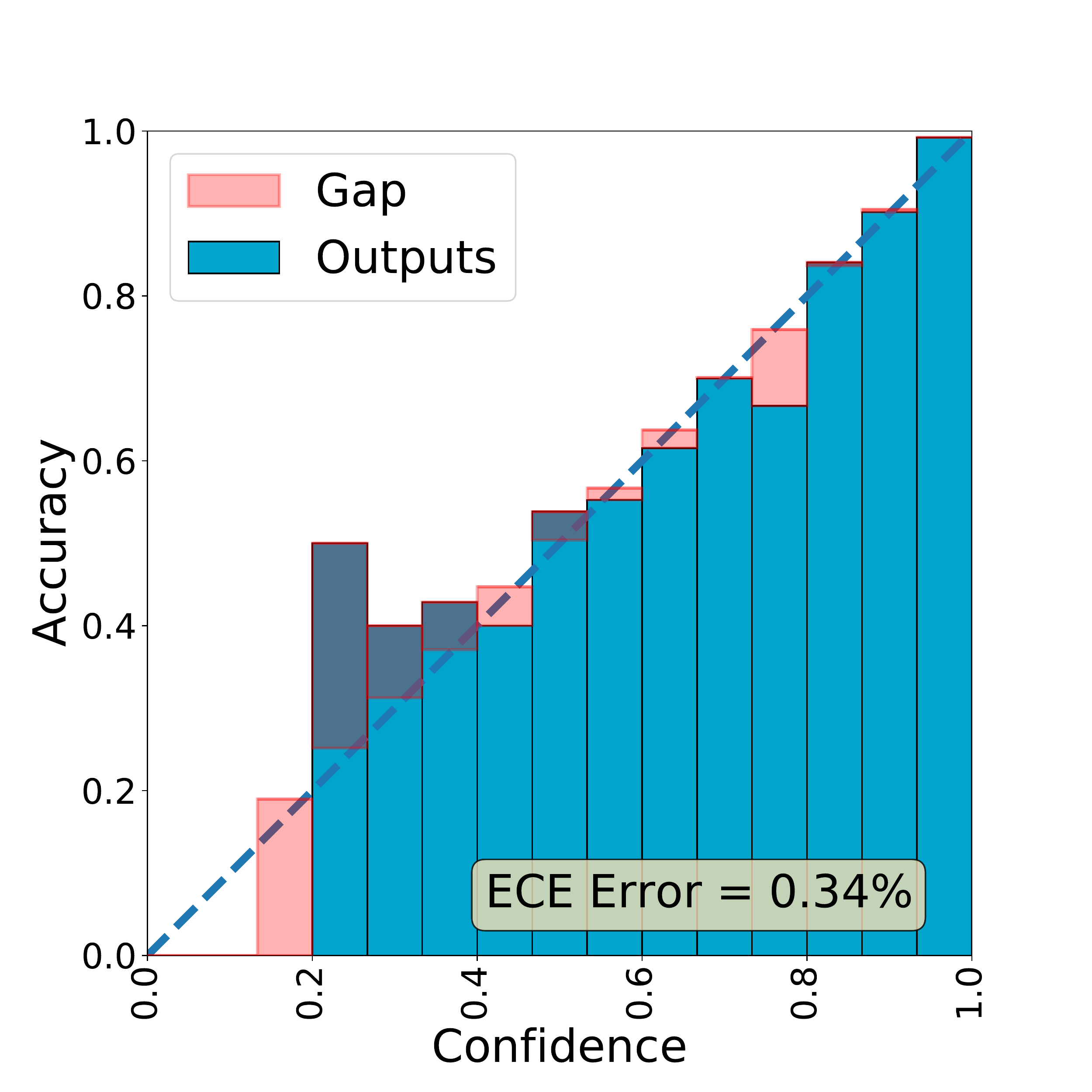}\\
(b) After calibration
\par\end{center}%
\end{minipage}
\par\end{center}%
\end{minipage}
\par\end{centering}
\caption{Reliability diagrams of the fine-tuned Inception-v3 network on the CUB dataset before and after calibration.}
\label{fig:callibration}
\end{figure}

We found that the pretrained classifier based on the Inception network (used to calculate IS) is uncalibrated or mis-calibrated. As a result, the classifier tends to be either over-confident or under-confident. This is verified by using expected calibration error (ECE) \cite{naeini2015obtaining} and reliability diagram \cite{degroot1983comparison,niculescu2005predicting}. ECE is the popular metric used to evaluate calibration whereas reliability diagram is a tool to visualize calibration quality. A classifier is well calibrated if they have a small ECE value and reliability diagram is close to identity. As can been seen in Figure~\ref{fig:callibration}(a), the Inception network, pretrained by StackGAN \cite{zhang2017stackgan} for evaluating recent text-to-image models on CUB, is under-confident. When computing the IS, this leads to inconsistency due to erroneous distance between conditional and marginal probability distributions. 

\setlength{\tabcolsep}{4pt}
\begin{table}
    \caption{A comparison between the IS and IS* scores on the CUB dataset. Thanks to the calibration step, our IS* no longer suffers the problem of counter models being ranked high despite producing bad results.}
	\begin{center}
		\begin{tabular}{l c c} 
			\toprule
			Method & IS ($\uparrow$) & IS* ($\uparrow$) ) \\
			\midrule
			GAN-INT-CLS~\cite{reed2016generative} & 2.73 & 7.51 \\
			
			StackGAN++~\cite{zhang2018stackgan++} & 4.10 & 12.69  \\
				
			AttnGAN~\cite{xu2018attngan} & 4.32 & 13.63  \\ 
			
			AttnGAN + CL~\cite{ye2021improving}  &  4.45  &  14.42  \\
			
			DM-GAN~\cite{zhu2019dm} & 4.68 & 15.00 \\
			
			DF-GAN~\cite{ming2020DFGAN} & 4.77  &  14.70  \\
			
			DM-GAN + CL~\cite{ye2021improving}  & 4.77  & \underline{15.08} \\
			
			\midrule
			Counter Model & \textbf{5.12} & 13.05  \\ 
			AttnGAN++ (ours) & \underline{4.78}  & \textbf{15.13}  \\
			\midrule
			\emph{Real Images} &  \emph{24.16}   &  \emph{46.27} \\
			\bottomrule
	\end{tabular}
\end{center}
\label{tab:ISstar_report}
\end{table}

To tackle this issue, we propose to calibrate the confidence score of the classifier, which we opt to apply the popular network calibration method of temperature scaling \cite{guo2017calibration}. Particularly, the classifier receives an input image $x$ and output a logit vector $z$. Before this logit vector $z$ is passed to a $\mathrm{softmax}$ layer to obtain probability values, we calibrate $z$ by scaling it with a positive scalar value $T$ for all classes. The conditional probability $p(y=k \mid x)^{*}$ with class label $k \in \{1..K\}$ after calibration is:
\begin{equation}
p(y=k \mid x)^* = \sigma(z/T)_{k},    
\end{equation}
where $K$ is the number of classes, $T$ is the temperature, and $\sigma$ represents the $\mathrm{softmax}$ function. We use the $p(y \mid x)^*$ vector for computing the divergence in IS*. The calibrated confidence score is $\max_k p(y=k \mid x)^*$. The value of $T$ is obtained by optimizing the negative log-likelihood loss on the validation set used to train the classifier. After calibration on CUB, we get $T = 0.598$. Figure~\ref{fig:callibration}(b) showed that after calibration, the under-confident issue is greatly mitigated illustrated by a significant drop in ECE error and a nearly diagonal shape of the plot. The IS* score shown in Table~\ref{tab:ISstar_report} demonstrated that the inconsistent score causing by the countermodel is also addressed by using IS* instead of IS. 

\noindent\textbf{\emph{Summary.}} Single-object text-to-image synthesis is a relatively well-explored topic. Challenges still arise with new tasks, e.g., validating the models with novel word compositions~\cite{park2021compositional}.  
Here we focused on the evaluation aspect and provided a unified benchmark with existing metrics and our IS* metric. Note that while both IS and FID are for image quality assessment, the benefit of IS (and our IS*) is that it does not require the distribution of real images for evaluation.

\section{Multiple-Object Text-to-Image Synthesis}
Evaluating text-to-image synthesis models with multiple objects is far more difficult than with a single object. The comprehensive survey by Frolov et al.~\cite{frolov2021adversarial} suggested many essential aspects for evaluating multiple-object text-to-image synthesis. We summarize these aspects in Table~\ref{tab:aspects}. As can be seen, simply using existing metrics as in the single-object case is insufficient because many critical aspects in the multi-object case have been implied or ignored, such as object count, relative position among objects, etc. In this section, we will describe a systematic approach for evaluating multi-object text-to-image models by revisiting and improving existing metrics and proposing new metrics for aspects that do not yet have a metric to quantify. Before we get into the specifics of the evaluation metrics, let us give an overview of the benchmark dataset that we use. Our benchmark is conducted on the MS-COCO version 2014 dataset \cite{lin2014microsoft}, which contains photos with many objects and complex backgrounds. We choose MS-COCO since this dataset is used popularly in developing text-to-image model with multiple objects. The setup for preparing training and validation set in our experiments are same with \cite{reed2016generative}. In particular, we employ the official training set of MS-COCO (approximately $80K$ images) as the training set of text-to-image models, and we test models on the MS-COCO validation set (approximately $40K$ images).

\subsection{Existing Metrics}  
\noindent\textbf{\emph{Image Realism.}} FID and our IS* can be used to analyze the photorealism of multi-object synthetic images in the same way they have been used for single object images. 

\noindent\textbf{\emph{Text Relevance.}} Current studies use RP to assess the alignment between text and the generated image. However, this metric is shown to overfit in multiple-object synthesis, having inconsistent ranking with real images, which can be seen in Figure~\ref{fig:teaser}. One reason for this is that previous works have used the same image and text encoders from DAMSM~\cite{xu2018attngan} for training and computing RP. To alleviate this overfitting issue, we use an independent text encoder and image encoder for RP. We selected CLIP~\cite{radford2021learning}, a powerful text and image encoders trained on a very large-scale dataset with $400$ million text-image pairs. This idea is also used by the concurrent work of Park et al.~\cite{park2021compositional}.
In our experiment, the overfitting problem of RP is mitigated using two new encoders, as demonstrated by the value of RP in real images have a large gap with the previous methods. A comparison between the traditional and our modified RP results can be found in supplementary material. 

\begin{table}[t]
\caption{Demanding aspects for the evaluation of multi-object text-to-image models presented by \cite{frolov2021adversarial} and our proposed metrics to assess the lacking criteria.}
\begin{center}
    \resizebox{\linewidth}{!}{%
	\begin{tabular}{l c c c c c c c c c} 
		\toprule
		Metric & Image & Object & Text & Object & Positional & Counting & Paraphrase  & Explainable & Automatic \\
		& Realism & Fidelity & Relevance & Accuracy & Alignment & Alignment & Robustness &  & \\
		\midrule
		IS~\cite{salimans2016improved} & \checkmark & & & & & & & & \checkmark \\
				
		FID~\cite{heusel2017gans} & \checkmark & & & & & & & & \checkmark \\
		
		RP~\cite{xu2018attngan} & & & \checkmark & & & & & & \checkmark \\
		
		SOA~\cite{hinz2019semantic} & & & \checkmark & \checkmark & & & & & \checkmark \\
		
		\midrule
		O-IS (Ours) & & \checkmark & & & & & & & \checkmark \\
		O-FID (Ours) & & \checkmark & & & & & & & \checkmark \\
		PA (Ours) & & & & & \checkmark & & & & \checkmark \\
		CA (Ours) & & & & & & \checkmark & & & \checkmark \\
		\midrule
		Human  & \checkmark & \checkmark & \checkmark & \checkmark & \checkmark & \checkmark & \checkmark & \checkmark &  \\
		\bottomrule
	\end{tabular}
	}
\end{center}
\label{tab:aspects}
\end{table}

\noindent\textbf{\emph{Object Accuracy.}} Semantic Object Accuracy (SOA)~\cite{hinz2019semantic} is proposed to measure whether generate images having the objects mentioned in the caption. Specifically, the authors proposed two sub-metrics including SOA-I (average recall between images) and SOA-C (average recall between classes), which are formulated as:
\begin{equation}
    \mathrm{SOA\mbox{-}C} = \dfrac{1}{|C|}\sum_{c \in C} \dfrac{1}{|I_{c}|}\sum_{i_{c} \in I_{c}} \mathrm{Object\mbox{-}Detector}(i_{c}), 
\end{equation}
\begin{equation}
    \mathrm{SOA\mbox{-}I} = \dfrac{1}{\sum_{c \in C}|I_{c}|}\sum_{c \in C}\sum_{i_{c}\in I_{c}}\mathrm{Object\mbox{-}Detector}(i_{c}),
\end{equation}
where $C$ is the object category set; $I_{c}$ is a set of images belonging to category c; $\mathrm{Object\mbox{-}Detector}(i_{c}) \in \{0, 1\}$ is an pretrained object detector returning 1 if the detector detect successfully an object belong to class $c$ in $i_{c}$. 

As can be seen, SOA is a plausible metric to evaluate the object accuracy factor in the text-to-image model. However, we found that both CPGAN~\cite{liang2019cpgan} and SOA used the same pre-trained YOLO-v3~\cite{redmon2018yolov3} in their implementation, which can potentially lead to overfitting. 
Empirically, the values of SOA-I and SOA-C of CPGAN are better than those for real images despite images from CPGAN are still non-realistic (Figure~\ref{fig:teaser}). 
To lessen the chance of overfitting, we choose Mask-RCNN~\cite{he2017mask} instead of YOLO-v3 to compute SOA.
The empirical result in our experiment shows that this selection helps mitigate the inconsistency problem. A comparison between the SOA results when using YOLO-v3 and Mask-RCNN can be found in the supplementary material. In this paper, we solely report SOA values computed by Mask-RCNN.

We now turn to describe our new metrics. As shown in Table~\ref{tab:aspects}, several aspects in evaluating multi-object text-to-image models remain lacking. Unsolved aspects that we will tackle in this paper include \textbf{\emph{Object Fidelity}}, \textbf{\emph{Positional Alignment}} and \textbf{\emph{Counting Alignment}}.
Positional alignment measures the relative position among the objects in the image, e.g., when there is a man and a tree in an image, whether `a man stands in front of a tree' and `a man stands behind a tree' affects the positional alignment. 
Counting alignment measures the compatibility of the number of objects illustrated by the input sentence and the generated image. Object fidelity evaluates the quality of the object set extracted from generated images.
In the survey by Frolov et al. \cite{frolov2021adversarial}, the authors simply provided a discussion without providing any concrete metrics for such aspects. In the following sections, we propose new metrics to address these shortcomings. 

\subsection{Object Fidelity}
Object-centric IS (O-IS) and Object-centric FID (O-FID) are our straightforward extensions of IS and FID with the aim to measure object fidelity in the generated images. In the literature, SceneFID~\cite{sylvain2021object} is the closest metric that can assess this criteria and is proposed for evaluating layout-to-image models. However, SceneFID requires the ground truth object bounding boxes from the layout to extract objects in the images preventing them to apply for the text-to-image task. In this work, we replace the need of using ground truth bounding boxes by leveraging the bounding boxes predicted by an off-the-shelf object detection model.  Specifically, we first use a well-trained object detector to localize and crop all object regions in each image in the generated image set. By treating all image regions as independently generated, we evaluate the fidelity by IS* and FID on the image regions, respectively.  
In our experiments, we used Mask-RCNN \cite{he2017mask} pre-trained on MS-COCO as the object detector. We also fine-tune and then calibrate the Inception-v3 classifier on the object dataset cropped from the images in MS-COCO based on ground truth bounding boxes to obtain a classifier having 80 classes, equaling the number of classes in MS-COCO. The Inception-v3 network after fine-tuning is used for both computing O-IS and O-FID.

\begin{figure}[t]
\centering
\includegraphics[width=0.76\linewidth]{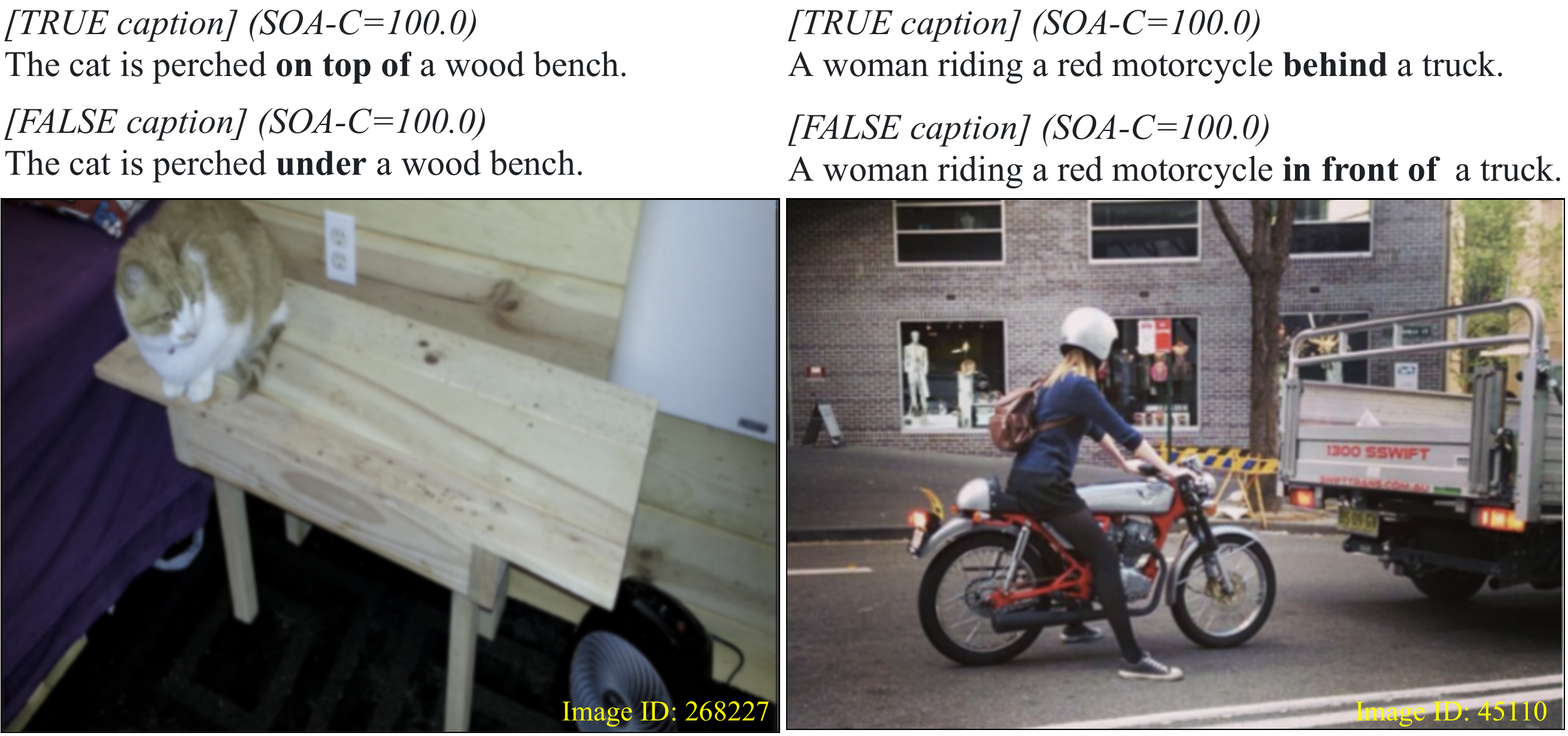}
\caption{Accessing positional alignment of the objects in the multi-object image is critical, yet it is still mostly ignored. This example shows a flaw in the existing metrics, such as SOA, which completely ignored the evaluation of positional alignment while maintaining good object accuracy. As can be seen, the SOA values for the image with the \emph{true} caption and with the \emph{false} caption are the same, which demonstrates that the SOA metric skips positional alignment. This weakness leads to the appearance of our Positional Alignment (PA) metric.
} 
\label{fig:soa_counter}
\end{figure}

\subsection{Positional Alignment}
Text descriptions are used to describe an image and typically include phrases that convey the positioning information between objects, such as \emph{behind, on top of, etc} (Figure~\ref{fig:soa_counter}). However, existing object-aware metrics like SOA do not penalize such incorrect relative object locations (e.g. generated images with inaccurate positional alignment still has high SOA scores). To tackle this issue, we propose a new metric to evaluate positional alignment, denoted by PA. 
First, we define the set of positional words as $W = $ \{\emph{above, right, far, outside, between, below, on top of, bottom, left, inside, in front of, behind, on, near, under}\}. 
For each word $w$ in $W$, we filter the captions having word $w$ in the evaluation set of the COCO dataset, and obtain the caption set $P_{w}$ for each word $w$. 
Each caption in $P_{w}$ is a matched caption, which means the image clearly explains the text. 
Given $P_{w}$, we build a mismatched caption by replacing $w$ in the matched caption by its antonym and keeping other words.
For example, the mismatched caption of "A man is \emph{in front of} the blue car" is "A man is \emph{behind} the blue car". 
Our evaluation begins by generating images from the matched captions in the test dataset.
For each word $w$ in $W$, we now have a set $D_{w}={\{(R_{wi}, P_{wi}, Q_{wi})\}}_{i=1}^{N_{w}}\}$ where $R_{wi}$ is a generated image from $P_{wi}$; $P_{wi}$ is matched caption; $Q_{wi}$ is mismatched caption of $P_{wi}$; $N_{w}$ is the number of captions having word $w$.
For each triplet in $D_{w}$, we use the image $R_{wi}$ to query the input caption from the binary query set including matched caption $P_{wi}$ and mismatched caption $Q_{wi}$. We mark a query as successful if the matched caption is successfully queried. The query success rate measures the positional alignment quality over all words:
\begin{align}
    \mathrm{PA}= \frac{1}{|W|} \sum_{w \in W} \frac{k_{w}}{N_{w}},
\end{align} 
where $k_{w}$ is the number of success cases, and $|W|$ is the total number of words. 
For image-to-text query, we use CLIP~\cite{radford2021learning} as our text-image matching model.

\subsection{Counting Alignment}
In the multi-object case, counting alignment is an vital factor but so far disregarded in current text-to-image synthesis evaluation. Therefore, we propose a metric for counting alignment (CA metric) that measures how closely the number of objects in a generated image matches the text description.

To evaluate with CA, we first need to construct the test data by filtering from captions in MS-COCO validation set the captions mentioned counting aspect such as \emph{a, one, two, three, four}. From these selected captions, we annotate the ground truth counting information for each one. It is worth noting that we only annotate the object types which can be counted by an object counter to avoid this metric to penalizing those object categories, which cannot be counted. For example, with a caption \emph{"A group of seven people having a light meal and discussion at a single large table"}, the ground truth counting is \emph{\{"person": 7.0, "dining table": 1.0\}}. Finally, we created a counting test set $D$ with $1000$ records. Each record has a form of $(t, c)$, in which $t$ is an input text description, and $c$ is the ground truth counting information.

We use a text-to-image model to generate images from each caption and use an off-the-shelf object counting model~\cite{cholakkal2019object} to count the number of objects for each object class from generated images. 
To get CA value, we compare the object count to the ground truth and measure the counting error using root mean squared error averaged over the test images:
\begin{equation}
    \mathrm{CA} = \frac{1}{|D|} \sum_{i=1}^{|D|} \sqrt{\dfrac{1}{N_{ic}}\sum_{j=1}^{N_{ic}}(\hat{c_{ij}} - c_{ij})^2},
\end{equation}
where $c_{ij}$ and $\hat{c_{ij}}$ is the ground truth and predicted object count in the image $i$ for object class $j$; $N_{ic}$ is the number of ground truth object types in image $i$, $|D|$ is the number of test samples.
    
\subsection{Ranking Score}
To facilitate the benchmark, we propose a simple formula to compute an average score for ranking purpose. The ranking score is calculated by summing all \emph{rankings} of the considered metrics. 
To the best of our knowledge, a similar approach is used in the nuScenes challenge for autonomous driving~\cite{caesar2020nuscenes} that ranks object detection methods by combining metrics for different bounding box properties such as center, orientation, and dimensions. In our case, since some evaluation aspects could have more than one metric variant, the ranking for each aspect is the average of the ranking of the variants. We treat all metrics and aspects equally, and thus use $\frac{1}{2}$ weight for IS and FID in image realism, O-IS and O-FID in object fidelity, SOA-I and SOA-C in object accuracy; other metrics have a unit weight. Our ranking score (RS) is computed as
\begin{align}
\mathrm{RS} &= \frac{1}{2}(\#\mathrm{IS^{*}} + \#\mathrm{FID}) + \frac{1}{2}(\#\mathrm{O\mbox{-}IS} + \#\mathrm{O\mbox{-}FID}) \\ \nonumber
&+ \frac{1}{2}(\#\mathrm{SOA\mbox{-}I} + \#\mathrm{SOA\mbox{-}C}) + \#\mathrm{PA} + \#\mathrm{CA} + \#\mathrm{RP},
\end{align}
where $\#(\mathrm{metric}) \in \{1..N\}$ denotes the ranking by a particular metric with $N$ is the number of considered methods.
 
\subsection{Benchmark Results}
\begin{table}[t]
\caption{Benchmark performances of the multi-object text-to-image synthesis models on the MS-COCO dataset. The \textbf{best} and \underline{runner-up} values are marked in bold and underline, respectively. As can be seen, our AttnGAN++ gains the competitive results compared to the current state-of-the-art text-to-image synthesis methods.
}
\begin{center}
    \resizebox{\linewidth}{!}{%
	\begin{tabular}{l c c c c c c c c c c c} 
	\toprule
			
	Method & IS* ($\uparrow$)  & FID ($\downarrow$) & RP($\uparrow$)  & SOA-C($\uparrow$) & SOA-I ($\uparrow$) & O-IS ($\uparrow$) & O-FID ($\downarrow$) & CA ($\downarrow$) & PA ($\uparrow$) & RS ($\uparrow$) \\
	
	\midrule
	GAN-CLS~\cite{reed2016generative} & 8.10 & 192.09 & 10.00 & 5.31 & 5.71 & 2.46 & 51.13 & 2.51  & 32.79  & 7.0 \\
	
	StackGAN~\cite{zhang2017stackgan} & 15.50 & 53.44 & 9.10 & 9.24 & 9.90 & 3.36 & 29.09 & 2.41 & 34.33 & 11.5 \\
	
	AttnGAN~\cite{xu2018attngan} & 33.79 & 36.90 & 50.56 & 47.13 & 49.78 & 5.04 & 20.92 & 1.82 & 40.08 & 29.0 \\ 
	
	DM-GAN~\cite{zhu2019dm} & 45.63 & 28.96 & 66.98 & 55.77 & 58.11 & 5.22 & 17.48 &  1.71 & 42.83 & 41.0 \\
	
	CPGAN~\cite{liang2019cpgan} & \textbf{59.64} & 50.68 & 69.08 & \textbf{81.86} & \textbf{83.83}  & \textbf{6.38} & 20.07 & 2.07 & 43.28 & 43.0  \\
	
	DF-GAN~\cite{ming2020DFGAN} & 30.45 & \textbf{21.05} & 42.44 & 37.85 & 40.19  & 5.12 & \textbf{14.39} & 1.96 & 40.39 & 31.5 \\
    
    AttnGAN + CL~\cite{ye2021improving} & 36.85 & 26.93 & 57.52 & 47.45 & 49.33  & 4.92 & 19.92 & 1.72 & 43.92 & 37.0 \\
    
    DM-GAN + CL~\cite{ye2021improving} & 46.61 & \underline{22.60} & \underline{70.36} & 58.68 & 61.05  & 5.09 & 15.50 & \underline{1.66} & \textbf{49.06} & \underline{51.5} \\
	
	DALLE-mini (zero-shot)~\cite{Dayma_DALLE_Mini_2021} & 19.82 & 62.90 & 48.72 & 26.64 & 27.90  & 4.10 & 23.83 & 2.31 & 47.39 & 23.5 \\
	
    \midrule
    
	AttnGAN++ (Ours) & \underline{54.63} & 26.58 & \textbf{72.48} & \underline{67.83} & \underline{69.97} & \underline{6.01} & \underline{15.43} & \textbf{1.57} & \underline{47.75} & \textbf{56.0} \\ 
	\midrule
	\emph{Real Images} & \emph{51.25} & \emph{2.62} & \emph{83.54} & \emph{90.02} & \emph{91.19} & \emph{8.63} & \emph{0.00} & \emph{1.05} & \emph{100.0} & \emph{65.0} \\
	
	\bottomrule
	
	\end{tabular}
	}
\end{center}
\label{tab:coco_benchmark}
\end{table}

We show the benchmark results in Table~\ref{tab:coco_benchmark}, from which we draw some following insights. 
Firstly, our proposed metrics (O-IS, O-FID, CA, PA) and two improved version of existing metrics (RP, SOA), properly rank real images as the best. An exception is IS* which ranks AttnGAN++ and CPGAN better than real images. However we opt to retain this metric due to its excellent properties on the single-object case, and the ranking score is consistent to human when including IS*. Second, our AttnGAN++ is ranked top for multi-object text-to-image synthesis in terms of overall performance, demonstrating that it is a substantial strong baseline for both single-object and multiple-object instances. Third, breaking down each part of our evaluation pipeline allows us to more clearly analyze each model's flaws and strengths than earlier evaluations. For examples, CPGAN outperforms other techniques on SOA-I and SOA-C since it explicitly considers object-level information in the training phase. DM-GAN + CL is the most effective method for positional alignment. While our AttnGAN++ performs better in the remaining aspects. The details of aspect's scores for each method are included in the supplementary material. 

\subsection{Human Evaluation} 

To ensure that our evaluations are reliable, we conducted a user analysis to test the metrics against assessments done by humans. We opt for 5 methods including StackGAN, AttnGAN, DM-GAN, CPGAN, AttnGAN++ (ours), and real images to conduct our user survey. We sample 50 test captions from MS-COCO and use the above methods to generate an image for each caption. The IDs for these captions are provided in supplementary for reproducibility. We ask each human subject ($40$ participants in total) to score each method from 1 (worst) to 5 (best) based on two criteria: \emph{plausibility} -- whether the image is plausible based on the content of the caption (object accuracy, counting, and positional alignment, text relevance), and \emph{naturalness} -- whether the image looks natural. 
The score of each human subject for each method is the sum of score of 50 images and divide by $250$ for normalization. 
The final score of each method is an average of the scores of each participant. 
Our evaluation result in Table~\ref{tab:user_study} shows that our final ranking is well-aligned with human evaluation.

\begin{table}[t]
\caption{Human evaluation results on the MS-COCO dataset. In this table, ranking scores (RS) are recalculated using just $5$ considered techniques and real photos. As can be observed, RS is well-aligned with human decisions.} 
\begin{center}
	\begin{tabular}{l c c} 
		\toprule
		Method & Ranking Score ($\uparrow$) &  Human Score ($\uparrow$) \\ 
		\midrule
		StackGAN~\cite{zhang2017stackgan} & 6.00 & 28.45 \\
		AttnGAN~\cite{xu2018attngan} & 13.5 & 37.40 \\
		DM-GAN~\cite{zhu2019dm}  &  20.0  & 41.47  \\
		CPGAN~\cite{liang2019cpgan} & 23.0 & 43.73 \\
		\midrule
		AttnGAN++ (ours)  &  \textbf{28.5}  & \textbf{45.01} \\
		\midrule
		\emph{Real Images} &  \emph{35.0}  &  \emph{99.82} \\
		\bottomrule
	\end{tabular}
\end{center}
\label{tab:user_study}
\end{table}

\section{Conclusion}
This paper performed an empirical study with benchmarks for text-to-image synthesis methods for both single-object and multiple-object scenario. The benchmark results reveal the inconsistency issues in the existing metrics, prompting us to propose the improved version of existing metrics as well as new metrics to evaluate many vital but lacking aspects in the multiple-object case. Our extensive experiments show that this bag of metrics provides a better and more consistent ranking with real images and human evaluation.

Our bag of metrics for text-to-image synthesis is by no means perfect. The proposed metrics can be further extended for complex cases, for example, to handle more positional words for positional alignment score and indefinite numeral adjectives (e.g., several, many) for counting alignment. 

\clearpage
\bibliographystyle{splncs04}
\bibliography{egbib}

\begin{thebibliography}{10}
\providecommand{\url}[1]{\texttt{#1}}
\providecommand{\urlprefix}{URL }
\providecommand{\doi}[1]{https://doi.org/#1}

\bibitem{brock2018large}
Brock, A., Donahue, J., Simonyan, K.: Large scale gan training for high
  fidelity natural image synthesis. arXiv preprint arXiv:1809.11096  (2018)

\bibitem{caesar2020nuscenes}
Caesar, H., Bankiti, V., Lang, A.H., Vora, S., Liong, V.E., Xu, Q., Krishnan,
  A., Pan, Y., Baldan, G., Beijbom, O.: nuscenes: A multimodal dataset for
  autonomous driving. In: CVPR (2020)

\bibitem{cholakkal2019object}
Cholakkal, H., Sun, G., Khan, F.S., Shao, L.: Object counting and instance
  segmentation with image-level supervision. In: CVPR (2019)

\bibitem{Dayma_DALLE_Mini_2021}
Dayma, B., Patil, S., Cuenca, P., Saifullah, K., Abraham, T., Le~Khac, P.,
  Melas, L., Ghosh, R.: Dall·e mini (2021),
  \url{https://github.com/borisdayma/dalle-mini}

\bibitem{degroot1983comparison}
DeGroot, M.H., Fienberg, S.E.: The comparison and evaluation of forecasters.
  Journal of the Royal Statistical Society: Series D (The Statistician)
  \textbf{32}(1-2) (1983)

\bibitem{ding2021cogview}
Ding, M., Yang, Z., Hong, W., Zheng, W., Zhou, C., Yin, D., Lin, J., Zou, X.,
  Shao, Z., Yang, H., et~al.: Cogview: Mastering text-to-image generation via
  transformers. NeurIPS  (2021)

\bibitem{frolov2021adversarial}
Frolov, S., Hinz, T., Raue, F., Hees, J., Dengel, A.: Adversarial text-to-image
  synthesis: A review. Neural Networks  (2021)

\bibitem{goodfellow2014generative}
Goodfellow, I., Pouget-Abadie, J., Mirza, M., Xu, B., Warde-Farley, D., Ozair,
  S., Courville, A., Bengio, Y.: Generative adversarial nets. In: NeurIPS
  (2014)

\bibitem{gulrajani2017improved}
Gulrajani, I., Ahmed, F., Arjovsky, M., Dumoulin, V., Courville, A.: Improved
  training of wasserstein gans. arXiv preprint arXiv:1704.00028  (2017)

\bibitem{guo2017calibration}
Guo, C., Pleiss, G., Sun, Y., Weinberger, K.Q.: On calibration of modern neural
  networks. arXiv preprint arXiv:1706.04599  (2017)

\bibitem{he2017mask}
He, K., Gkioxari, G., Doll{\'a}r, P., Girshick, R.: Mask r-cnn. In: ICCV (2017)

\bibitem{heusel2017gans}
Heusel, M., Ramsauer, H., Unterthiner, T., Nessler, B., Hochreiter, S.: Gans
  trained by a two time-scale update rule converge to a local nash equilibrium.
  In: NeurIPS (2017)

\bibitem{hinz2019semantic}
Hinz, T., Heinrich, S., Wermter, S.: Semantic object accuracy for generative
  text-to-image synthesis. arXiv preprint arXiv:1910.13321  (2019)

\bibitem{hong2018inferring}
Hong, S., Yang, D., Choi, J., Lee, H.: Inferring semantic layout for
  hierarchical text-to-image synthesis. In: CVPR (2018)

\bibitem{ioffe2015batch}
Ioffe, S., Szegedy, C.: Batch normalization: Accelerating deep network training
  by reducing internal covariate shift. arXiv preprint arXiv:1502.03167  (2015)

\bibitem{kang2020contragan}
Kang, M., Park, J.: Contragan: Contrastive learning for conditional image
  generation  (2020)

\bibitem{karnewar2020msg}
Karnewar, A., Wang, O.: Msg-gan: Multi-scale gradients for generative
  adversarial networks. In: CVPR (2020)

\bibitem{karras2017progressive}
Karras, T., Aila, T., Laine, S., Lehtinen, J.: Progressive growing of gans for
  improved quality, stability, and variation. arXiv preprint arXiv:1710.10196
  (2017)

\bibitem{karras2019style}
Karras, T., Laine, S., Aila, T.: A style-based generator architecture for
  generative adversarial networks. In: CVPR (2019)

\bibitem{karras2020analyzing}
Karras, T., Laine, S., Aittala, M., Hellsten, J., Lehtinen, J., Aila, T.:
  Analyzing and improving the image quality of stylegan. In: CVPR (2020)

\bibitem{kodali2017convergence}
Kodali, N., Abernethy, J., Hays, J., Kira, Z.: On convergence and stability of
  gans. arXiv preprint arXiv:1705.07215  (2017)

\bibitem{le2015tiny}
Le, Y., Yang, X.: Tiny imagenet visual recognition challenge. CS 231N  (2015)

\bibitem{li2019control}
Li, B., Qi, X., Lukasiewicz, T., H.~S.~Torr, P.: Controllable text-to-image
  generation. arXiv preprint arXiv:1909.07083  (2019)

\bibitem{li2019object}
Li, W., Zhang, P., Zhang, L., Huang, Q., He, X., Lyu, S., Gao, J.:
  Object-driven text-to-image synthesis via adversarial training. In: CVPR
  (2019)

\bibitem{liang2019cpgan}
Liang, J., Pei, W., Lu, F.: Cpgan: Full-spectrum content-parsing generative
  adversarial networks for text-to-image synthesis. arXiv preprint
  arXiv:1912.08562  (2019)

\bibitem{lim2017geometric}
Lim, J.H., Ye, J.C.: Geometric gan. arXiv preprint arXiv:1705.02894  (2017)

\bibitem{lin2014microsoft}
Lin, T.Y., Maire, M., Belongie, S., Hays, J., Perona, P., Ramanan, D.,
  Doll{\'a}r, P., Zitnick, C.L.: Microsoft coco: Common objects in context. In:
  ECCV (2014)

\bibitem{maaten2008visualizing}
Maaten, L.v.d., Hinton, G.: Visualizing data using t-sne. Journal of Machine
  Learning Research  \textbf{9}(Nov) (2008)

\bibitem{mirza2014conditional}
Mirza, M., Osindero, S.: Conditional generative adversarial nets. arXiv
  preprint arXiv:1411.1784  (2014)

\bibitem{miyato2018spectral}
Miyato, T., Kataoka, T., Koyama, M., Yoshida, Y.: Spectral normalization for
  generative adversarial networks. arXiv preprint arXiv:1802.05957  (2018)

\bibitem{naeini2015obtaining}
Naeini, M.P., Cooper, G.F., Hauskrecht, M.: Obtaining well calibrated
  probabilities using bayesian binning. In: AAAI (2015)

\bibitem{niculescu2005predicting}
Niculescu-Mizil, A., Caruana, R.: Predicting good probabilities with supervised
  learning. In: ICML (2005)

\bibitem{Nilsback08}
Nilsback, M.E., Zisserman, A.: Automated flower classification over a large
  number of classes. In: Proceedings of the Indian Conference on Computer
  Vision, Graphics and Image Processing (Dec 2008)

\bibitem{odena2016conditional}
Odena, A., Olah, C., Shlens, J.: Conditional image synthesis with auxiliary
  classifier gans. ICML  (2017)

\bibitem{park2021compositional}
Park, D.H., Azadi, S., Liu, X., Darrell, T., Rohrbach, A.: Benchmark for
  compositional text-to-image synthesis. In: NeurIPS Datasets and Benchmarks
  Track (2021)

\bibitem{qiao2019mirrorgan}
Qiao, T., Zhang, J., Xu, D., Tao, D.: Mirrorgan: Learning text-to-image
  generation by redescription. In: CVPR (2019)

\bibitem{radford2021learning}
Radford, A., Kim, J.W., Hallacy, C., Ramesh, A., Goh, G., Agarwal, S., Sastry,
  G., Askell, A., Mishkin, P., Clark, J., et~al.: Learning transferable visual
  models from natural language supervision. arXiv preprint arXiv:2103.00020
  (2021)

\bibitem{radford2015unsupervised}
Radford, A., Metz, L., Chintala, S.: Unsupervised representation learning with
  deep convolutional generative adversarial networks. arXiv preprint
  arXiv:1511.06434  (2015)

\bibitem{pmlr-v139-ramesh21a}
Ramesh, A., Pavlov, M., Goh, G., Gray, S., Voss, C., Radford, A., Chen, M.,
  Sutskever, I.: Zero-shot text-to-image generation. In: ICML (2021)

\bibitem{razavi2019generating}
Razavi, A., van~den Oord, A., Vinyals, O.: Generating diverse high-fidelity
  images with vq-vae-2. In: NeurIPS (2019)

\bibitem{redmon2018yolov3}
Redmon, J., Farhadi, A.: Yolov3: An incremental improvement. arXiv preprint
  arXiv:1804.02767  (2018)

\bibitem{reed2016generative}
Reed, S., Akata, Z., Yan, X., Logeswaran, L., Schiele, B., Lee, H.: Generative
  adversarial text-to-image synthesis. In: ICML (2016)

\bibitem{russakovsky2015imagenet}
Russakovsky, O., Deng, J., Su, H., Krause, J., Satheesh, S., Ma, S., Huang, Z.,
  Karpathy, A., Khosla, A., Bernstein, M., et~al.: Imagenet large scale visual
  recognition challenge. IJCV  \textbf{115}(3) (2015)

\bibitem{salimans2016improved}
Salimans, T., Goodfellow, I., Zaremba, W., Cheung, V., Radford, A., Chen, X.:
  Improved techniques for training gans. In: NeurIPS (2016)

\bibitem{shen2020interpreting}
Shen, Y., Gu, J., Tang, X., Zhou, B.: Interpreting the latent space of gans for
  semantic face editing. In: CVPR (2020)

\bibitem{shen2020interfacegan}
Shen, Y., Yang, C., Tang, X., Zhou, B.: Interfacegan: Interpreting the
  disentangled face representation learned by gans. arXiv preprint
  arXiv:2005.09635  (2020)

\bibitem{sylvain2021object}
Sylvain, T., Zhang, P., Bengio, Y., Hjelm, R.D., Sharma, S.: Object-centric
  image generation from layouts. In: AAAI (2021)

\bibitem{szegedy2016rethinking}
Szegedy, C., Vanhoucke, V., Ioffe, S., Shlens, J., Wojna, Z.: Rethinking the
  inception architecture for computer vision. In: CVPR (2016)

\bibitem{tan2019semantics}
Tan, H., Liu, X., Li, X., Zhang, Y., Yin, B.: Semantics-enhanced adversarial
  nets for text-to-image synthesis. In: ICCV (2019)

\bibitem{ming2020DFGAN}
Tao, M., Tang, H., Wu, F., Jing, X.Y., Bao, B.K., Xu, C.: Df-gan: A simple and
  effective baseline for text-to-image synthesis. In: CVPR (2022)

\bibitem{theis2015note}
Theis, L., Oord, A.v.d., Bethge, M.: A note on the evaluation of generative
  models. arXiv preprint arXiv:1511.01844  (2015)

\bibitem{vaswani2017attention}
Vaswani, A., Shazeer, N., Parmar, N., Uszkoreit, J., Jones, L., Gomez, A.N.,
  Kaiser, {\L}., Polosukhin, I.: Attention is all you need. In: NeurIPS (2017)

\bibitem{welinder2010caltech}
Welinder, P., Branson, S., Mita, T., Wah, C., Schroff, F., Belongie, S.,
  Perona, P.: Caltech-ucsd birds 200. Tech. Rep. CNS-TR-2010-001, California
  Institute of Technology (2010)

\bibitem{wu2019logan}
Wu, Y., Donahue, J., Balduzzi, D., Simonyan, K., Lillicrap, T.: Logan: Latent
  optimisation for generative adversarial networks. arXiv preprint
  arXiv:1912.00953  (2019)

\bibitem{xu2018attngan}
Xu, T., Zhang, P., Huang, Q., Zhang, H., Gan, Z., Huang, X., He, X.: Attngan:
  Fine-grained text to image generation with attentional generative adversarial
  networks. In: CVPR (2018)

\bibitem{ye2021improving}
Ye, H., Yang, X., Takac, M., Sunderraman, R., Ji, S.: Improving text-to-image
  synthesis using contrastive learning. arXiv preprint arXiv:2107.02423  (2021)

\bibitem{yin2019semantics}
Yin, G., Liu, B., Sheng, L., Yu, N., Wang, X., Shao, J.: Semantics
  disentangling for text-to-image generation. In: CVPR (2019)

\bibitem{zhang2019self}
Zhang, H., Goodfellow, I., Metaxas, D., Odena, A.: Self-attention generative
  adversarial networks. In: ICML (2019)

\bibitem{zhang2021cross}
Zhang, H., Koh, J.Y., Baldridge, J., Lee, H., Yang, Y.: Cross-modal contrastive
  learning for text-to-image generation (2021)

\bibitem{zhang2017stackgan}
Zhang, H., Xu, T., Li, H., Zhang, S., Wang, X., Huang, X., Metaxas, D.N.:
  Stackgan: Text to photo-realistic image synthesis with stacked generative
  adversarial networks. In: ICCV (2017)

\bibitem{zhang2018stackgan++}
Zhang, H., Xu, T., Li, H., Zhang, S., Wang, X., Huang, X., Metaxas, D.N.:
  Stackgan++: Realistic image synthesis with stacked generative adversarial
  networks. TPAMI  \textbf{41}(8) (2018)

\bibitem{zhu2019dm}
Zhu, M., Pan, P., Chen, W., Yang, Y.: Dm-gan: Dynamic memory generative
  adversarial networks for text-to-image synthesis. In: CVPR (2019)

\end{thebibliography}

\clearpage
\appendix
\noindent\textbf{\LARGE Supplementary Material}\par\bigskip
\noindent This supplemental document provides more details of our bag of metrics. We first present the details of our improved IS* metric including the impact of the calibration step on the IS* score (Section~\ref{sec:calib}) and the implementation details for the counter model for reproducing the inconsistency problem of IS (Section~\ref{sec:counter}). We then detailed our improved versions of RP and SOA and show how our metrics can mitigate the overfitting issues in the original versions in the multi-object text-to-image synthesis (Section~\ref{sec:rp_soa}). Next, we detail our benchmark including the statistics of our test data (Section~\ref{sec:stats}), a complete benchmark of multi-object text-to-image models based on each assessment criterion (Section~\ref{sec:multi_benchmark}), the architecture and network configurations of our AttnGAN++ baseline (Section~\ref{sec:attnganpp}) as well as more visual examples (Section~\ref{sec:visual}) and t-SNE visualization (Section~\ref{sec:tsne}). Finally, we provide the caption ids we used in our user study (Section~\ref{sec:user_study_details}).

\section{Details for our IS* metric}

\subsection{Impact of calibration on the classifier and IS*}
\label{sec:calib}

\begin{figure}
\begin{centering}
\noindent\begin{minipage}[t]{1\columnwidth}%
\begin{center}
\begin{minipage}[t]{0.45\columnwidth}%
\begin{center}
\includegraphics[width=0.9\columnwidth,height=0.9\columnwidth]{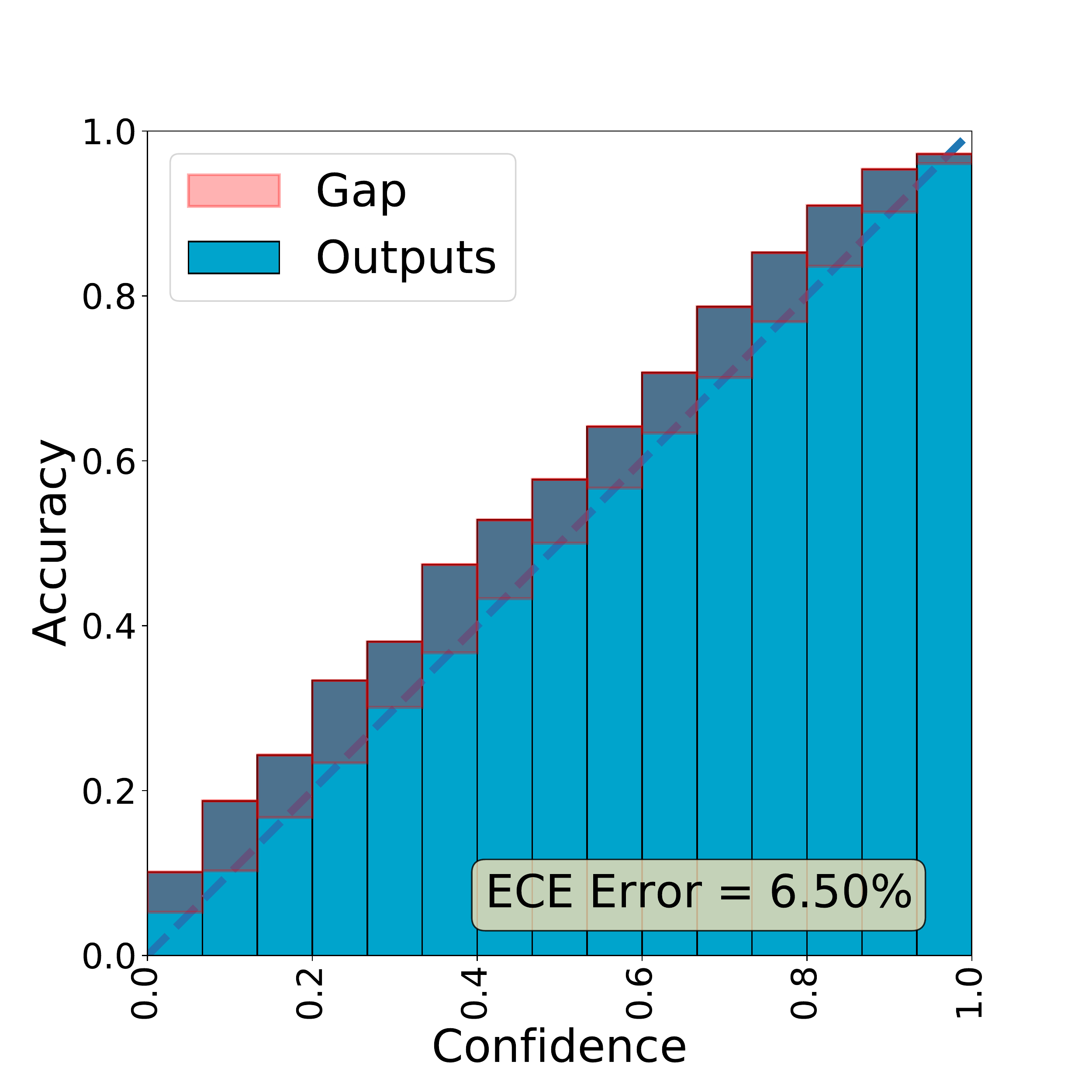}\\
(a) Before calibration
\par\end{center}%
\end{minipage}\quad{}%
\begin{minipage}[t]{0.45\columnwidth}%
\begin{center}
\includegraphics[width=0.9\columnwidth,height=0.9\columnwidth]{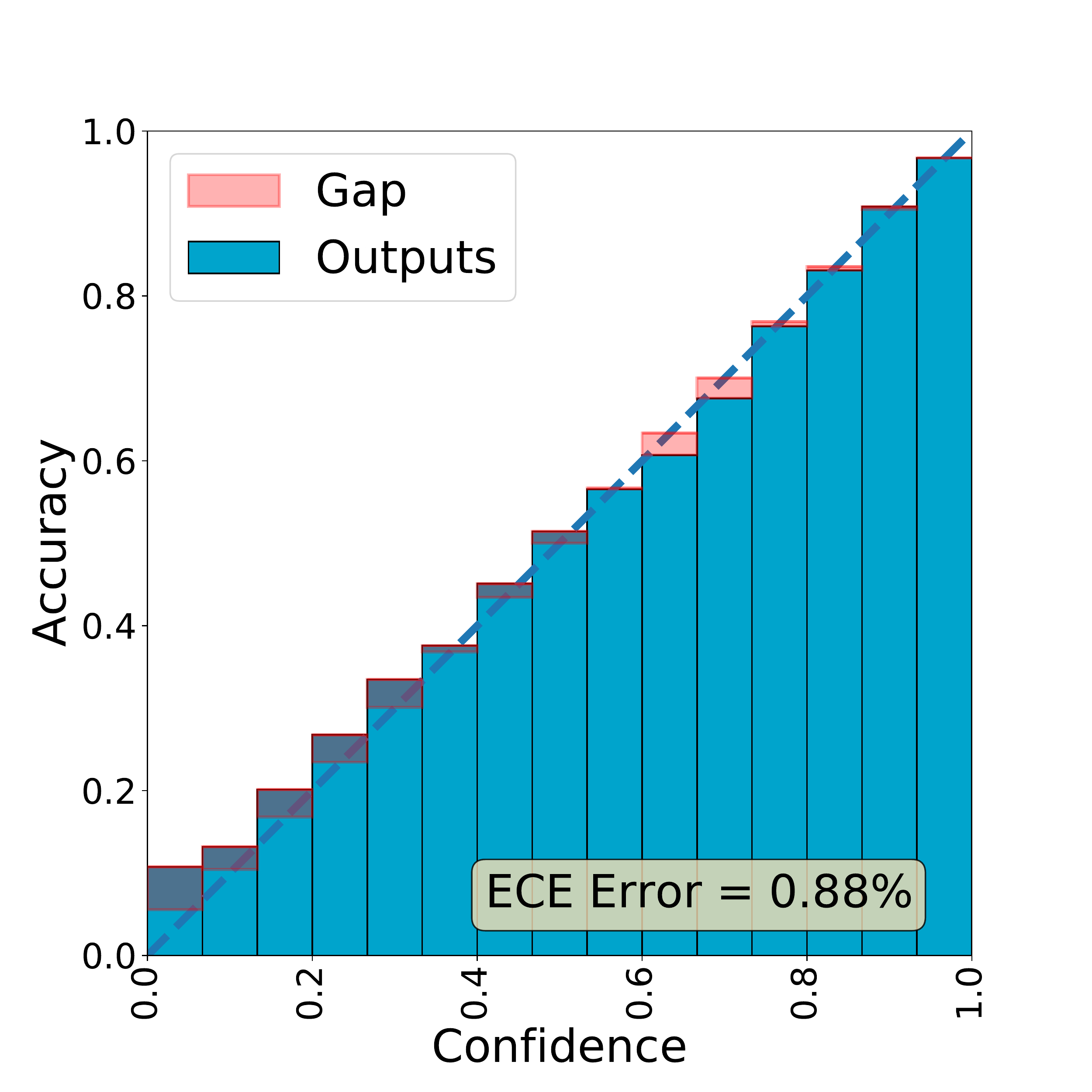}\\
(b) After calibration
\par\end{center}%
\end{minipage}
\par\end{center}%
\end{minipage}
\par\end{centering}
\caption{Reliability diagrams of the Inception-v3 network pre-trained on the ImageNet dataset before and after calibration.}
\label{fig:1000_classes_callibration}
\end{figure}

In the main paper, we showed that severe miscalibration of the classifier used to compute IS on the CUB dataset in previous methods led to inconsistent IS scores of a counter model, and we proposed IS* to fix this issue.  In this section, we further conduct another experiment to verify the impact of the calibration step causing on computing IS*. In detail, this experiment is performed on the vanilla GAN task, which is the Tiny ImageNet~\cite{le2015tiny} image generation. The classifier used to measure IS in these works is the Inception-v3 network pre-trained on ImageNet~\cite{russakovsky2015imagenet}. It is worth noting that this classifier is used popularly for measuring IS in the traditional GAN image generation task. The IS and IS* results are shown in Table~\ref{tab:gan_tiny_image_net}. We also plot the reliability diagrams and ECE errors of this classifier before and after calibration in Figure~\ref{fig:1000_classes_callibration}. As can be seen, even before calibration, this classifier is noticeably well calibrated. Hence, the effect of the calibration process on this classifier is negligible demonstrated through the temperature $T$ after calibration is $0.909$ ($T=1$ means calibration does not have any effects on classifier). Therefore, we would only see the local ranking differs in IS and IS*. 

\begin{table}
\label{tab:gan_tiny_image_net}
\caption{Comparing the ranking of IS and IS* on Tiny ImageNet dataset with various GAN models. The cases, which are ranked inconsistently between IS and IS*, are marked in \textbf{bold}. As we expect, only local ranking differs between IS and IS* appear due to the well-calibrated of the classifier even before calibration.}
\begin{center}
	\begin{tabular}{ l  c c } 
		\toprule
		Method & IS ($\uparrow$) & IS* ($\uparrow$) \\
		\midrule
		WGAN-GP~\cite{gulrajani2017improved}  & 1.64 & 1.79  \\
 		GGAN~\cite{lim2017geometric}  & 5.22 & 7.00  \\
		DCGAN~\cite{radford2015unsupervised}  & 5.70 & 7.79 \\
		ACGAN~\cite{odena2016conditional} & 6.51 & 9.30  \\
		BigGAN-LO~\cite{wu2019logan} & 7.83 & 11.29  \\
		SNGAN~\cite{miyato2018spectral} & \textbf{8.38} & \textbf{12.36} \\
		SAGAN~\cite{zhang2019self}  & \textbf{8.48} & \textbf{12.34} \\
		WGAN-DRA~\cite{kodali2017convergence} & 9.35 & 14.00 \\
		BigGAN~\cite{brock2018large}  & 12.43 & 18.80 \\
		ContraGAN~\cite{kang2020contragan}  & 13.76 & 21.64 \\
		\bottomrule
	\end{tabular}
\end{center}
\end{table}

\subsection{Counter model implementation}
\label{sec:counter}
\begin{figure}
    \begin{center}
      \includegraphics[width=0.8\linewidth]{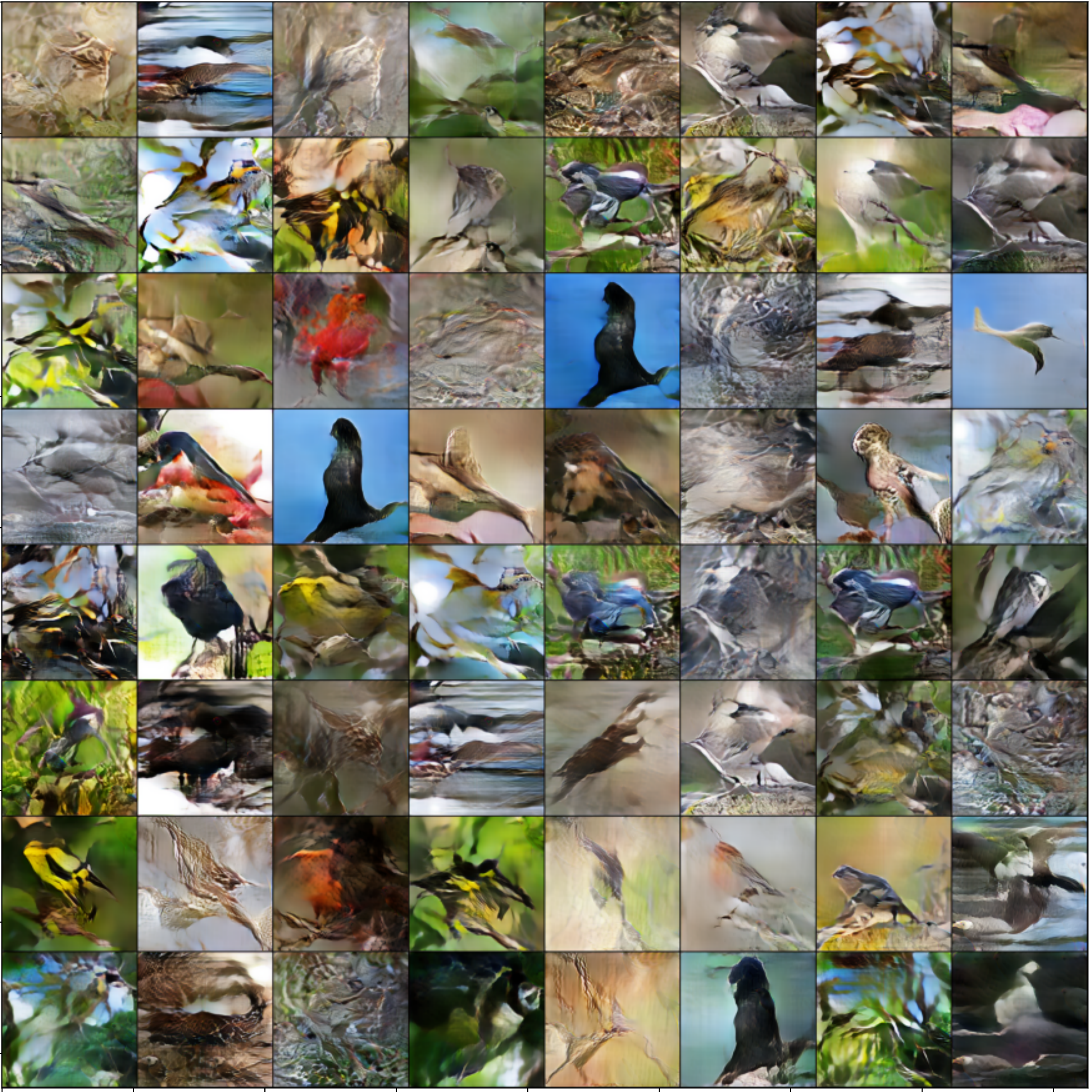}
    \end{center}
    \caption{More random (not curated) visual samples from our countermodel on the CUB dataset. As can be seen, the synthesized images are not realistic in most cases.}
    \label{fig:counter_example_more_examples}
\end{figure}

This section details the development of our counter model, which was utilized to demonstrate the inconsistency problem of IS. Our counter model is built on AttnGAN++ and MSG-GAN~\cite{karnewar2020msg}. Table~\ref{tab:counter_model_settings} shows the network details of the counter model. The training and evaluation configurations can be found in Table~\ref{tab:training_settings}. More random (not curated) visual samples synthesized by the counter model are also provided in Figure~\ref{fig:counter_example_more_examples} to demonstrate that these samples from the counter model are quite poor in comparison to those from AttnGAN++.

\section{Details of our improved RP and SOA}
\label{sec:rp_soa}
In the main paper, we presented that the existing versions of RP and SOA overfit in the multi-object scenario of the MS-COCO dataset, as evidenced by the fact that the values from some methods on these metrics exceed the corresponding values from real photos although these methods produce images with poorer quality than real photos.  We demonstrate this by comparing the values of the original and our modified versions of these metrics in Table~\ref{tab:r_precision_comparison} and Table~\ref{tab:soa_comparison}. As can be seen, the overfitting phenomena on RP and SOA is fully eliminated in our enhanced versions.

\begin{table}[t]
\begin{center}
\caption{Comparison of the original version of RP and our improved version one on the MS-COCO~\cite{lin2014microsoft} dataset. Inconsistent results are marked in \textbf{bold}. As can be observed, the value of RP on real photographs is the lowest, showing the original RP's heavily overfitting issue. Noticeably, our improved RP alleviated significantly by using CLIP~\cite{radford2021learning}. Note that the values of RP in these experiments are calculated from $30,000$ captions as most of the previous works do. In the main paper, we only sample $5,000$ captions and calculate RP from these captions to save time but guarantee consistent scores.}
\label{tab:r_precision_comparison}
\resizebox{0.8\linewidth}{!}{%
	\begin{tabular}{ l  c  c } 
		\toprule
		Method & R-precision (original) ($\uparrow$) & R-precision (ours) ($\uparrow$)\\
		\midrule
		StackGAN~\cite{zhang2017stackgan} & \textbf{72.03} & 38.46 \\
	
		AttnGAN~\cite{xu2018attngan} & \textbf{83.76} & 50.92 \\ 
		
		DM-GAN~\cite{zhu2019dm} & \textbf{92.23} &  65.91 \\
		
		CPGAN~\cite{liang2019cpgan} & \textbf{93.59} & 70.36 \\
		
		\midrule
		
		AttnGAN++ (ours) & \textbf{96.39} & 73.37 \\
		
		\midrule
		
		Real Images & 67.35 & 83.65 \\
		
		\bottomrule
	\end{tabular}%
}
\end{center}
\end{table}

\begin{table}
\caption{Comparison of the original version of SOA (including SOA-I and SOA-C) and our improved version of SOA on the MS-COCO~\cite{lin2014microsoft} dataset. Inconsistent results are highlighted in \textbf{bold}, which shows that SOA-C and SOA-I of CPGAN are higher than real images and our SOA greatly migrated this phenomenon. Note that the values of SOA in this experiment are calculated on full captions provided by the authors of this metric, while the ones we report by our TISE toolbox are computed on the sample from them (about $16k$ captions) to save time but output the consistent scores.}
\label{tab:soa_comparison}
\begin{center}
\resizebox{0.8\linewidth}{!}{%
	\begin{tabular}{l c c c c} 
		\toprule
		Method & SOA-C ($\uparrow$) & SOA-I ($\uparrow$) & SOA-C ($\uparrow$) & SOA-I ($\uparrow$)\\
		& (original) & (original) & (ours) & (ours) \\
		\midrule
		StackGAN~\cite{zhang2017stackgan} & 21.09 & 30.35 & 31.34 & 49.97 \\
	
		AttnGAN~\cite{xu2018attngan} & 25.88 & 39.01 & 47.26 & 62.02 \\ 
		
		DM-GAN~\cite{zhu2019dm} & 33.44 & 48.03 & 55.40 & 68.76 \\
		
		CPGAN~\cite{liang2019cpgan} & \textbf{77.02} & \textbf{84.55} & 82.25 & 88.97 \\
		
		\midrule
		
		AttnGAN++(ours) & 48.33 & 67.19 & 67.52 & 76.33 \\
		
		\midrule
		
		Real Images & 74.97 & 80.84 & 89.98 & 92.92 \\
		
		\bottomrule
	\end{tabular}%
	}
\end{center}
\end{table}

\section{Details of our benchmark}

\subsection{Benchmark data}
\label{sec:stats}
In the previous works, the inconsistency in the construction of testing data has caused many difficulties in benchmark models. A comprehensive survey~\cite{frolov2021adversarial} also pointed that there are some metrics are reported with inconsistent numbers between different research works. We find out that the non-unified input test data is one of the reasons leading to this issue. Therefore, we provide unified testing data in our TISE toolbox in order to compare techniques fairly. 

\begin{figure}[h!]
\begin{center}
   \includegraphics[width=0.6\linewidth]{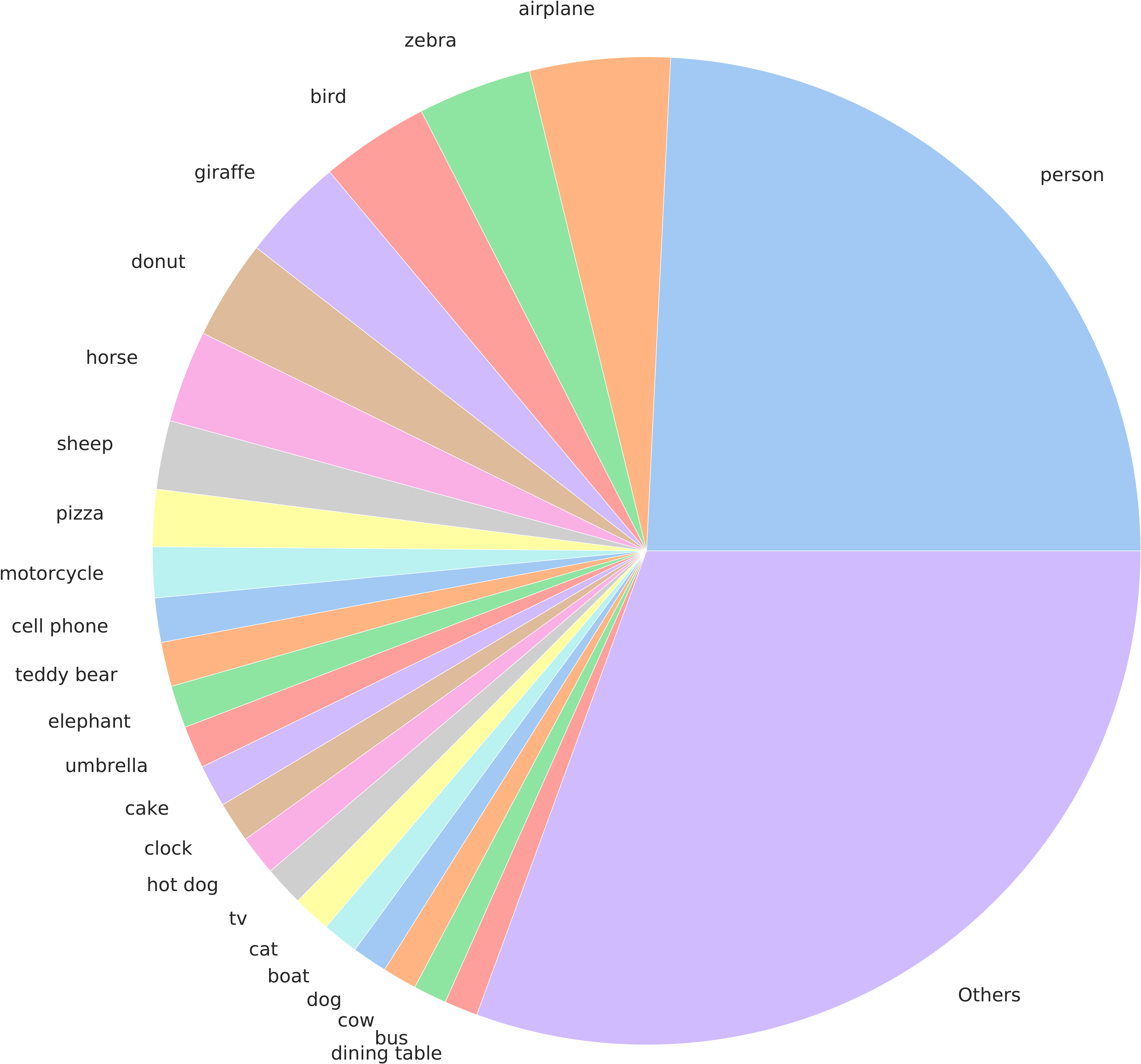}
\end{center}
\caption{Distribution of the number of object classes in our provided testing data for counting alignment factor in multi-object case. Best viewed in zoom.}
\label{fig:counting_data_statistic}
\end{figure}

\begin{figure}[h!]
\begin{center}
\includegraphics[width=0.6\linewidth]{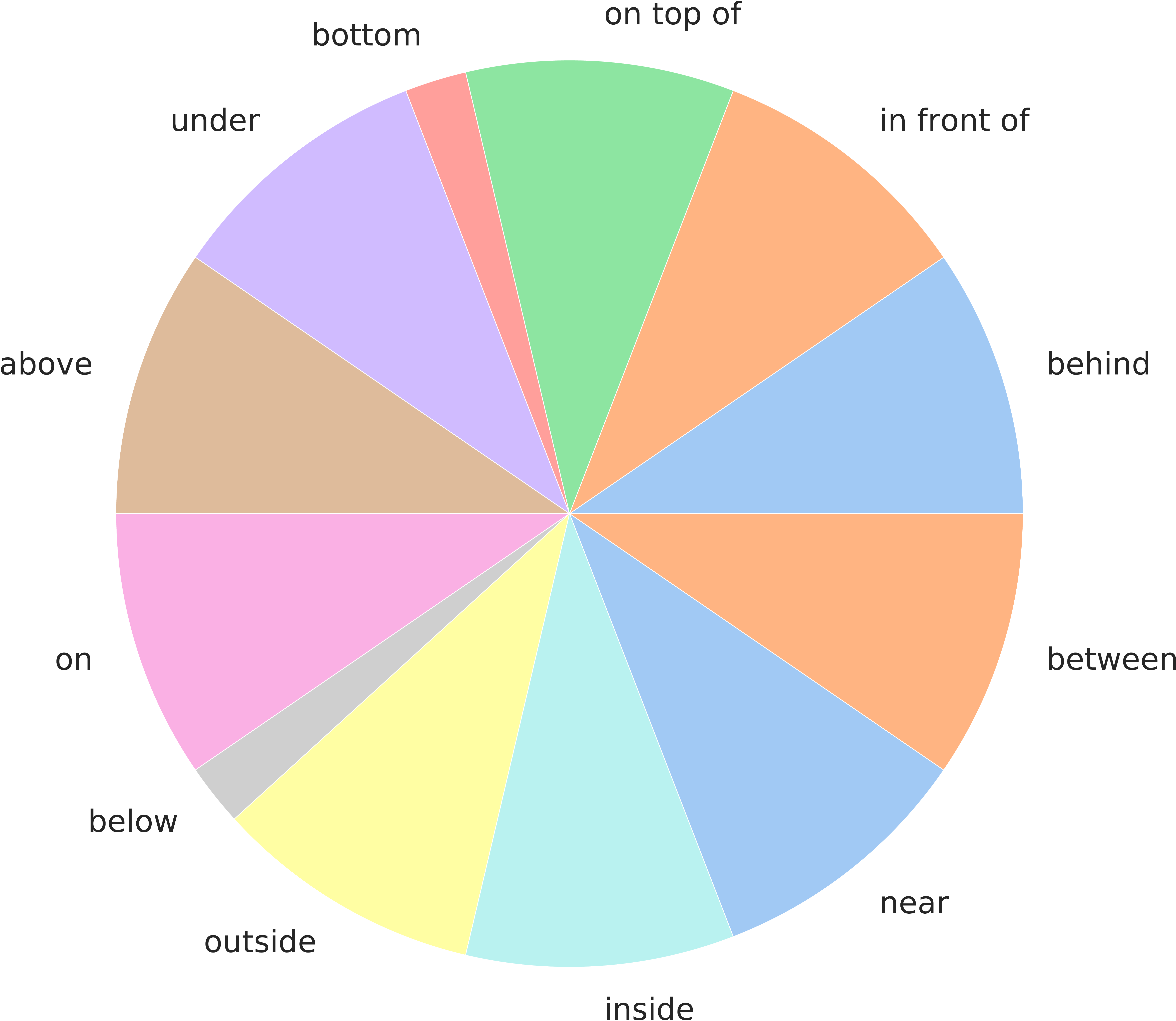}
\end{center}
\caption{Distribution of the number of test captions having corresponding positional words considered in our PA metric.}
\label{fig:pa_data_statistic}
\end{figure}

The details of our test data is as follows.  The number of captions used in each metrics are shown in Table~\ref{tab:cub_num_captions}  and Table~\ref{tab:coco_num_captions} for CUB and MS-COCO, respectively. The distribution of per-class  object count and positional words for counting alignment (CA) and positional alignment (PA) metric are visualized in Figure~\ref{fig:counting_data_statistic} and Figure~\ref{fig:pa_data_statistic}, respectively.

\begin{table}[h]
\caption{The number of test captions used in evaluation on the CUB dataset.}
\label{tab:cub_num_captions}
\begin{center}
\resizebox{.45\linewidth}{!}{%
	\begin{tabular}{ l c } 
		\toprule
		Metric & \#Captions\\
		\midrule
		Image Realism (IS, FID) & $30,000$ \\
		Text Relevance (RP) & $30,000$ \\
		\bottomrule
	\end{tabular}%
}
\end{center}
\end{table}

\begin{table}[h]
\caption{The number of test captions used in evaluating each evaluation aspect on the MSCOCO dataset.}
\label{tab:coco_num_captions}
\begin{center}
\resizebox{.5\linewidth}{!}{%
	\begin{tabular}{ l  c } 
		\toprule
		Metric & \#Captions\\
		\midrule
		Image Realism (IS, FID) & $10,000$ \\
		Object Fidelity (O-IS, O-FID) & $10,000$ \\ 
		Text Relevance (RP) & $5,000$ \\
		Object Accuracy (SOA-C,  SOA-I) & $15,223$\\
		Positional Alignment (PA) & $1,046$ \\
		Counting Alignment (CA) & $1,000$ \\ 
		\bottomrule
	\end{tabular}%
}
\end{center}
\end{table}

\clearpage
\subsection{Benchmark on MS-COCO for each evaluation aspect}
\label{sec:multi_benchmark}
Table~\ref{tab:aspect_scores} provides the details of aspect's ranking scores for each method on MS-COCO dataset. Here we show the performance of each model on six evaluation criteria, including Image Realism, Object Accuracy, Text Relevance, Object Accuracy, Object Fidelity, Counting Alignment, Positional Alignment.

\begin{table}[t]
\caption{The details of the ranking scores for each evaluation aspect of each method on the MS-COCO~\cite{lin2014microsoft} dataset. \textbf{Best} and \underline{runner-up} values are marked in bold and underline, respectively.}
\label{tab:aspect_scores}
\begin{center}
\resizebox{\linewidth}{!}{%
\begin{tabular}{l c c c c c c} 
\toprule
Method & Image & Text & Object & Object & Counting  & Positional \\
        & Realism & Relevance & Accuracy & Fidelity & Alignment & Alignment \\
\midrule
GAN-CLS~\cite{reed2016generative} & 1.0 & 2.0 & 1.0 & 1.0 & 1.0 & 1.0 \\

StackGAN~\cite{zhang2017stackgan} & 2.5 & 1.0 & 2.0 & 2.0 & 2.0 & 2.0 \\

AttnGAN~\cite{xu2018attngan} & 5.0 & 5.0 & 5.5 & 4.5 & 6.0 & 3.0 \\

DM-GAN~\cite{zhu2019dm} & 6.5 & 7.0 & 7.0 & 7.5 & 8.0 & 5.0 \\

CPGAN~\cite{liang2019cpgan} & 7.5 & 8.0 & \textbf{10.0} & 7.5 & 4.0 & 6.0 \\

DF-GAN~\cite{ming2020DFGAN} & 7.0 & 3.0 & 4.0 & \underline{8.5} & 5.0 & 4.0 \\

AttnGAN + CL~\cite{ye2021improving} & 6.5 & 6.0 & 5.5 & 5.0 & 7.0 & 7.0 \\

DM-GAN + CL~\cite{ye2021improving} & \underline{8.5} & \underline{9.0} & 8.0 & 7.0 & \underline{9.0} & \textbf{10.0} \\

DALLE-mini (zero-shot)~\cite{Dayma_DALLE_Mini_2021} & 2.5 & 4.0 & 3.0 & 3.0 & 3.0 & 8.0 \\

\midrule

AttnGAN++ (Ours) & \textbf{9.0} & \textbf{10.0} & \underline{9.0} & \textbf{9.0} & \textbf{10.0} & \underline{9.0} \\
\midrule
\emph{Real Images} & \emph{10.0} & \emph{11.0} & \emph{11.0} & \emph{11.0} & \emph{11.0} & \emph{11.0} \\
\bottomrule
\end{tabular}
}
\end{center}
\end{table}

\section{AttnGAN++ Architecture} 
\label{sec:attnganpp}
\begin{figure*}[t]
    \begin{center}
       \includegraphics[width=\linewidth,height=0.68\linewidth]{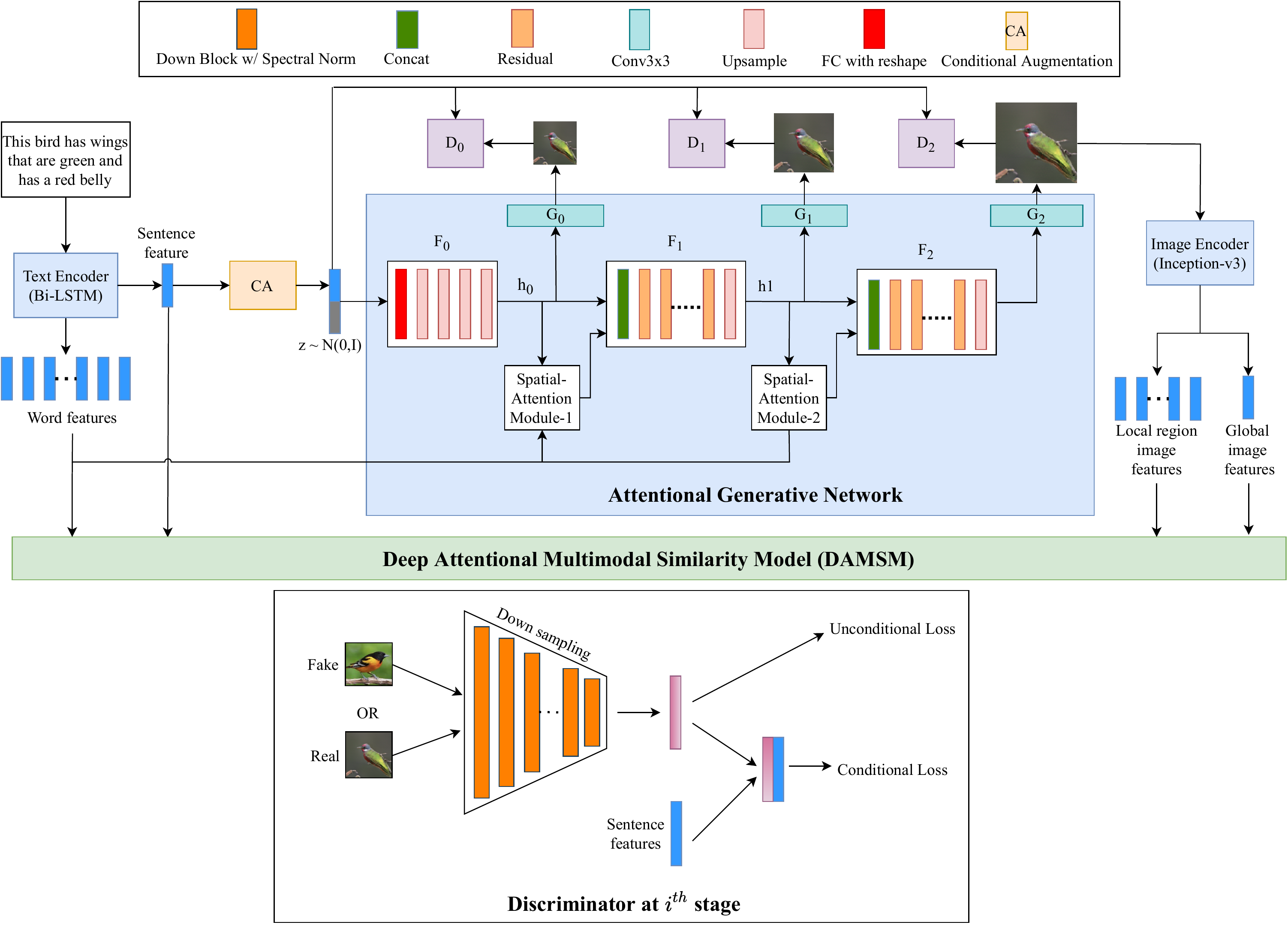}
    \end{center}
    \caption{The architecture of our AttnGAN++. In each discriminator, we employ spectral normalization~\cite{miyato2018spectral} to each convolution layer instead of using batch normalization~\cite{ioffe2015batch} as in the original AttnGAN~\cite{xu2018attngan}. Implementation details for each layer can be found in Table~\ref{tab:attn_gan_plus_plus_settings}.}
    \label{fig:attngan_architecture}
\end{figure*}

Along with the assessment toolkit, we also offered our AttnGAN++, a new baseline based on AttnGAN~\cite{xu2018attngan}. The main difference between AttnGAN++ and AttnGAN is that we apply spectral normalization~\cite{miyato2018spectral} to discriminators to stabilize the training process of GAN. With this simple technique, the performance of the model is boosted significantly comparing with the original version. The architecture of AttnGAN++ is shown in Figure~\ref{fig:attngan_architecture}. The network details and training settings of AttnGAN++ are demonstrated in Table~\ref{tab:attn_gan_plus_plus_settings} and Table~\ref{tab:training_settings} respectively. 

\section{More visual results}
\label{sec:visual}
\begin{figure}[t]
    \begin{center}
      \includegraphics[width=0.9\linewidth]{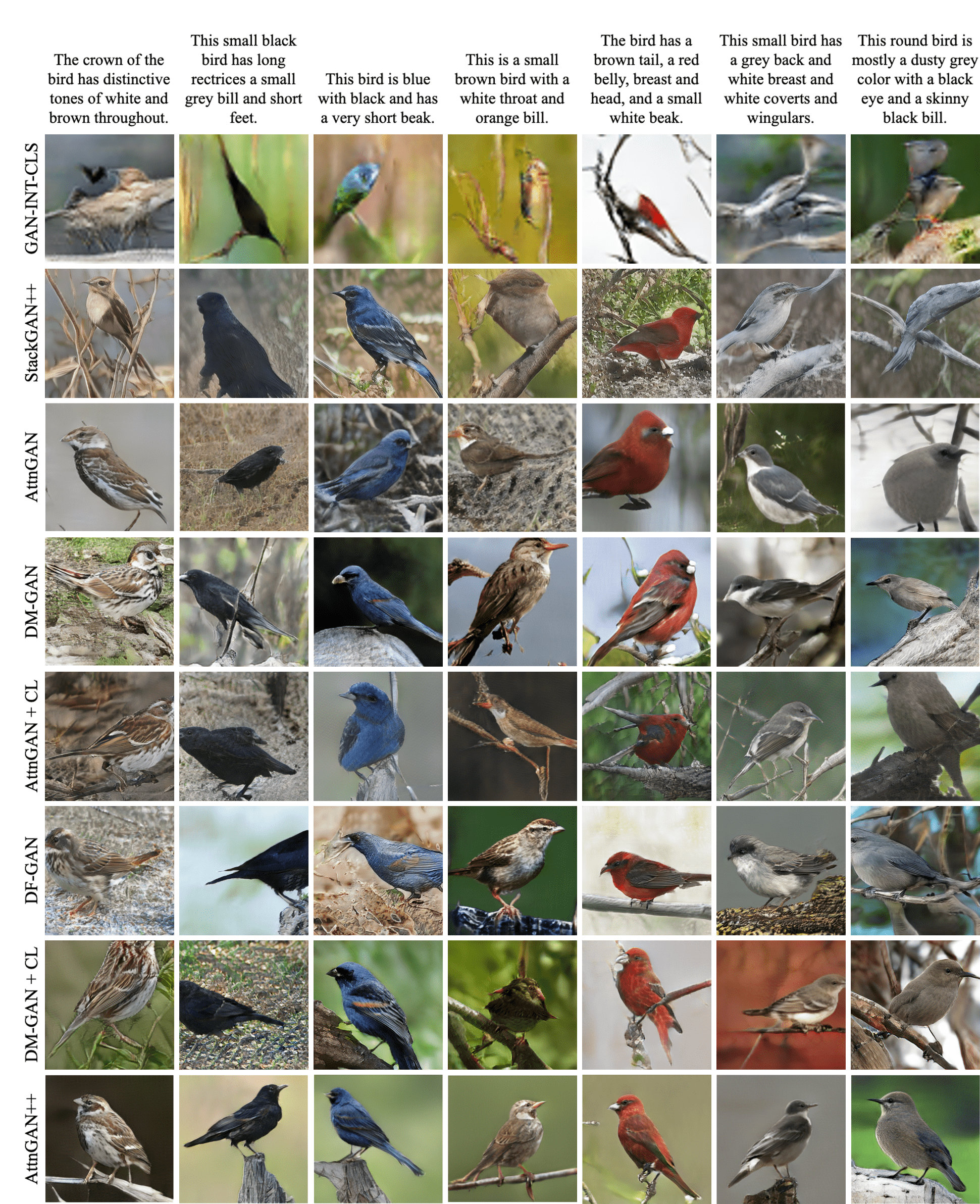}
    \end{center}
    \caption{Qualitative examples of the single object text-to-image generation models on the CUB~\cite{welinder2010caltech} dataset. Best viewed in zoom.}
    \label{fig:cub_qualitative_supp}
\end{figure}

\begin{figure}[t]
    \begin{center}
      \includegraphics[width=0.8\linewidth]{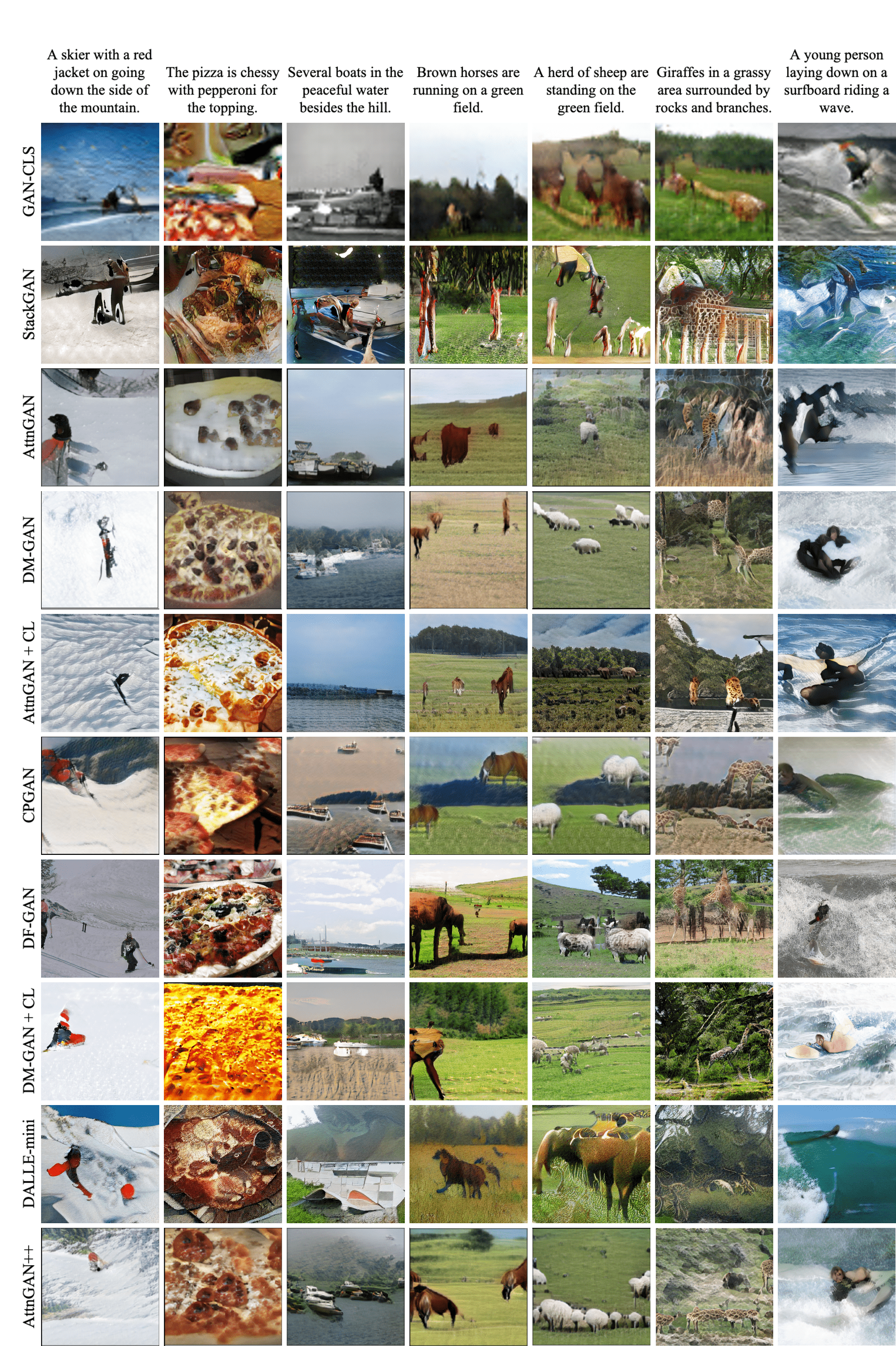}
    \end{center}
    \caption{Qualitative examples of the multi-object text-to-image synthesis models on the MS-COCO~\cite{lin2014microsoft} dataset. Best viewed in zoom.}
    \label{fig:coco_qualitative_supp}
\end{figure}

Additionally, we show more visual examples of our AttnGAN++ comparing with the current state-of-the-art text-to-image models on CUB~\cite{welinder2010caltech} dataset in Figure~\ref{fig:cub_qualitative_supp} and MS-COCO~\cite{lin2014microsoft} dataset in Figure~\ref{fig:coco_qualitative_supp} for qualitative measuring. 

\section{t-SNE Visualizations} 
\label{sec:tsne}
\begin{figure}
    \begin{center}
      \includegraphics[width=0.8\linewidth]{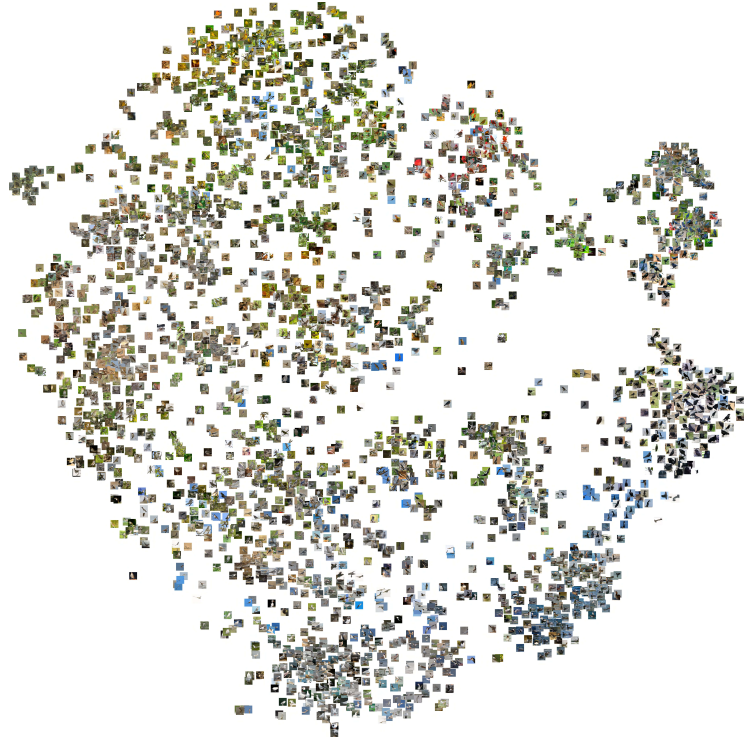}
    \end{center}
    \caption{Visualization of generated images from captions on the test set of CUB~\cite{welinder2010caltech} dataset by \textbf{our AttnGAN++} using t-SNE~\cite{maaten2008visualizing}. The number of clusters in the visualization shows that the photos generated by our model span a wide range of bird species. As a result of the similar appearances of various bird species, some clusters are near together and overlap slightly. We also found no intra-class mode dropping, indicating that the model does not create the same sample in each bird class over and over. As can be seen in each cluster, the samples are belonging to one bird class with a variety of poses, and backgrounds. Best viewed in zoom.}
    \label{fig:tnse_attngan_plus_plus}
\end{figure}

\begin{figure}
    \begin{center}
      \includegraphics[width=0.8\linewidth]{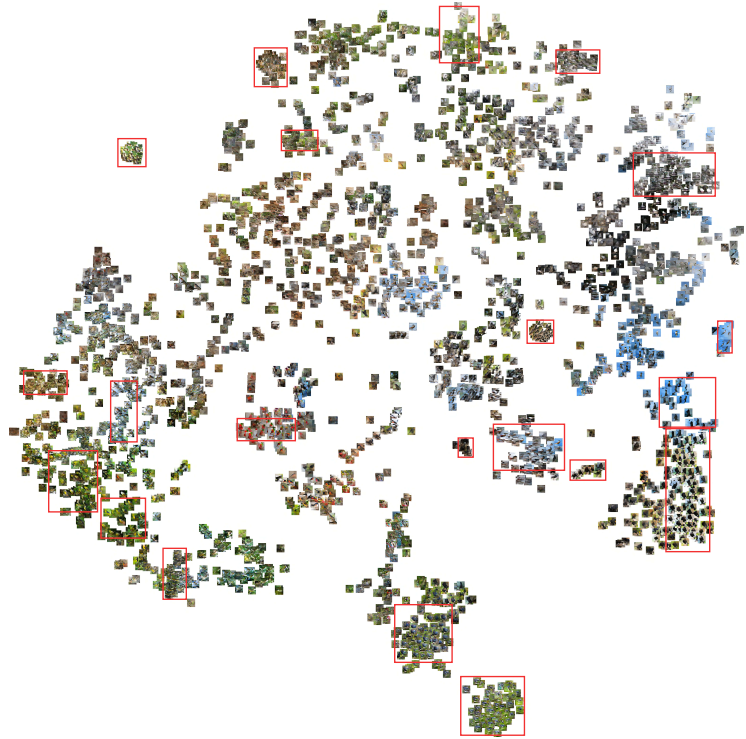}
    \end{center}
    \caption{Visualization of generated images from captions on the test set of CUB~\cite{welinder2010caltech} dataset by \textbf{counter model} using t-SNE~\cite{maaten2008visualizing}. As can be seen, the images of counterexample are not realistic. The counter model tends to generate only one sample again and again per class that are surrounded by red squares in the visualization. Best viewed in zoom.}
    \label{fig:tnse_counterexamples}
\end{figure}

\begin{figure}
    \begin{center}
      \includegraphics[width=0.8\linewidth]{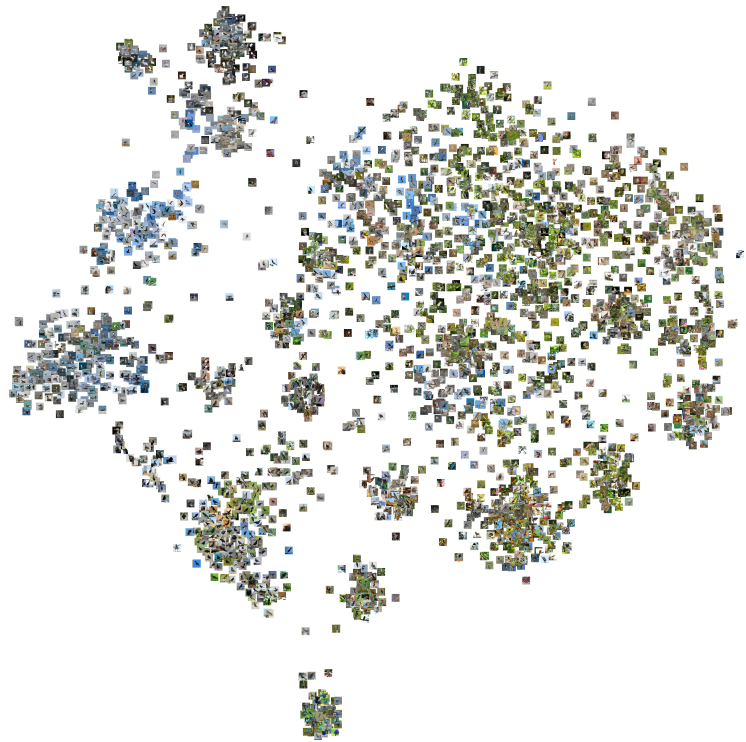}
    \end{center}
    \caption{Visualization of \textbf{real images} from CUB~\cite{welinder2010caltech} test set by using t-SNE~\cite{maaten2008visualizing}. Best viewed in zoom.}
    \label{fig:tnse_cub_real_images}
\end{figure}

To visualize the statistics of synthesized images, we utilize t-SNE~\cite{maaten2008visualizing}. Firstly, we extract feature vectors from these synthesized images using a pre-trained image encoder~\cite{xu2018attngan}. Then, we use t-SNE to convert these high dimensional feature vectors to 2-dimensional positions at which we display the images.  The t-SNE visualization of generated images by AttnGAN++ and counter model on CUB can be found in Figure~\ref{fig:tnse_attngan_plus_plus} and Figure~\ref{fig:tnse_counterexamples}, respectively. Additionally, we also show the t-SNE of all real images of the CUB test set in Figure~\ref{fig:tnse_cub_real_images} for reference. Note that the t-SNE image has a very high resolution so it is best viewed with an offline pdf viewer.

\section{User Study}
\label{sec:user_study_details}
To facilitate reproducibility, we provide the IDs of captions which we used in our human evaluation: \emph{503647, 302716, 817708, 72017, 563987, 434439, 375212, 478341, 737362, 323692, 177535, 338067, 810717, 416305, 680452, 439866, 558122, 545601, 196294, 380857, 782291, 324845, 767124, 63597, 648878, 73383, 327849, 799148, 829090, 107333, 805428, 371195, 443142, 394904, 754057, 421896, 361352, 517666, 75305, 625131, 202787, 723526, 569736, 442834, 183253, 642468, 277787, 150568, 502193, 643215}. 

\clearpage

\begin{table}[t]
\caption{Network details of our AttnGAN++. Some components which are not mentioned here such as text encoder, image encoder, DAMSM, its settings can be found in AttnGAN~\cite{xu2018attngan}. In the tables, k = kernel size, s = stride, p = padding, b = bias.}
\label{tab:attn_gan_plus_plus_settings}
\begin{centering}
\noindent\begin{minipage}[t]{1\columnwidth}%
\begin{center}
\begin{minipage}[t]{0.45\columnwidth}%
\begin{center}
\label{sub_tab:attn_gan_plus_plus_generator}
(a) Generator 
\resizebox{\columnwidth}{!}{%
\begin{tabular}{l | c} 
	\toprule
	Module & Output shape / Details \\
	\midrule
	\textbf{Up Block} & \\
	\emph{Params:} (in\_planes, out\_planes) & \\
	Input shape  & in\_planes $\times$ h $\times$ w \\
	Upsampling & Nearest Neighbor, scale factor $= 2$ \\
	Conv(k=3, s=1, p=1, b=False) & $2~*$ out\_planes $\times$ $h * 2$ $\times$ $w * 2$ \\ 
	BatchNorm2D & No change shape\\
	Gated Linear Unit (GLU) & out\_planes $\times$ $h*2$ $\times$ $w*2$ \\ 
    \\
	\textbf{Residual Block} & \\
	Input X & \\
	Conv(k=3, s=1, p=1, b=False) & Up channel size by 2, \\
	BatchNorm2D & No change shape\\
	Gated Linear Unit (GLU) & Down channel size by 2 \\
	Conv(k=3, s=1, p=1, b=False) & No change shape \\
	BatchNorm2D & No change shape\\
	Add output w/ X (skip connection) & No change shape \\
	\\
	\textbf{Spatial Attention layer} & see AttnGAN \\
	\\
	\textbf{Conditional Augumentation (CA)} & see AttnGAN \\
	\\
	\textbf{Generator $64 \times 64$} & \\
	\textit{Input} & \\
	Input noise & $nzf$\\
	Caption Embedding  & $nef$ \\
	\textit{Computation} & \\
	CA on caption embedding to get c & $ncf$ \\
	Concat c w/ noise & $ncf + nzf$ \\ 
	Linear(b=False) &  $ngf * 16 * 4 * 4 * 2$\\
	BatchNorm1D & No change shape \\
	Gated Linear Unit (GLU) & $ngf * 16 * 4 * 4 * 1$ \\
	Reshape & $16 * ngf \times 4 \times 4$ \\ 
	Up Block 1 & $8 * ngf \times 8 \times 8$ \\ 
	Up Block 2 & $4 * ngf \times 16 \times 16$ \\
	Up Block 3 & $2 * ngf \times 32 \times 32$ \\
	Up Block 4 & $ngf \times 64 \times 64$ \\
	Conv(k=3, s=1, p=1, b=False) & $3 \times 64 \times 64$ \\
	Tanh & No change shape \\
	\\
	\textbf{Generator $128 \times 128$} & \\
	\textit{Input} & \\
	Previous hidden features & $ngf \times 64 \times 64 $ \\
	Word Mask & $word\_num$ \\
	Word features & $nef \times word\_num$ \\ 
	\textit{Computation} & \\
	Spatial Attention Layer & $ngf \times 64 \times 64 $ \\
	Residual Block $\times$ residual\_num & $ngf \times 64 \times 64$ \\
	Concat w/ previous hidden features & $2 * ngf \times 64 \times 64$ \\
	Up Block & $ngf \times 128 \times 128$ \\
	Conv(k=3, s=1, p=1, b=False) & $3 \times 128 \times 128$ \\
	Tanh & No change shape \\
	\\
	\textbf{Generator $256 \times 256$} & \\
	\textit{Input} & \\
	Previous hidden features & $ngf \times 128 \times 128 $ \\
	Word Mask & $word\_num$ \\
	Word features & $nef \times word\_num$ \\ 
	\textit{Computation} & \\
	Spatial Attention Layer & $ngf \times 128 \times 128 $ \\
	Residual Block $\times$ residual\_num & $ngf \times 128 \times 128$ \\
	Concat w/ previous hidden features & $2 * ngf \times 128 \times 128$ \\
	Up Block & $ngf \times 256 \times 256$ \\
	Conv(k=3, s=1, p=1, b=False) & $3 \times 256 \times 256$ \\
	Tanh & No change shape \\
	\bottomrule
\end{tabular}%
}
\par\end{center}%
\end{minipage}\quad{}%
\hfill
\begin{minipage}[t]{0.45\columnwidth}%
\begin{center}
\label{sub_tab:attn_gan_plus_plus_discriminator}
(b) Discriminator 
\resizebox{\columnwidth}{!}{%
\begin{tabular}{l | c} 
	\toprule
	Module & Output shape / Details \\
	\midrule
	\textbf{Down Block} & \\
	\emph{Params:} (in\_planes, out\_planes) & \\
    Input shape  & in\_planes $\times$ h $\times$ w \\
	SpectralNorm(Conv(k=4, s=2, p=1, b=True) & out\_planes $\times$ h/2 $\times$ w/2 \\
	LeakyReLU(alpha=0.2) & No change shape\\
	\\
	\textbf{Block3x3\_leakyReLU} & \\
	\emph{Params:} (in\_planes, out\_planes) & \\
	Input shape  & in\_planes $\times$ h $\times$ w \\
	SpectralNorm(Conv(k=3, s=1, p=1, b=True) & out\_planes $\times$ h $\times$ w \\
	LeakyReLU(alpha=0.2) & No change shape \\
	\\
	\textbf{Discriminator $256 \times 256$} & \\
	Input tensor & $3 \times 256 \times 256$ \\
	Down Block & $ndf \times 128 \times 128$ \\
	Down Block & $2 * ndf \times 64 \times 64$ \\
	Down Block & $4 * ndf \times 32 \times 32$\\
	Down Block & $8 * ndf \times 16 \times 16$\\
	Down Block & $16 * ndf \times 8 \times 8$\\
	Down Block & $32 * ndf \times 4 \times 4$\\
	Block3x3\_leakyReLU & $16 * ndf \times 4 \times 4$\\
	Block3x3\_leakyReLU & $8 * ndf \times 4 \times 4$ \\
	\textit{Unconditional logits} & \\  
	Conv(k=4, s=4, p=0, b=True) & $1$ \\
	Sigmoid & $1$ \\ 
	\textit{Conditional logits} & \\ 
	Caption Embedding & $nef$\\
	Concat w/ replicated caption embedding & $8*ndf+nef \times 4 \times 4$ \\
	Block3x3\_leakyReLU & $8*ndf \times 4 \times 4$\\
	Conv(k=4, s=4, p=0, b=True) & $1$\\
	Sigmoid & $1$ \\
	\\
	\textbf{Discriminator $128 \times 128$} & \\
	Input tensor & $3 \times 128 \times 128$ \\
	Down Block & $ndf \times 64 \times 64$ \\
	Down Block & $2*ndf \times 32 \times 32$ \\
	Down Block & $4*ndf \times 16 \times 16$\\
	Down Block & $8*ndf \times 8 \times 8$\\
	Down Block & $16*ndf \times 4 \times 4$\\
	Block3x3\_leakyReLU & $8*ndf \times 4 \times 4$ \\
	\textit{Unconditional logits} & \\  
	Conv(k=4, s=4, p=0, b=True) & $1$ \\
	Sigmoid & $1$ \\ 
	\textit{Conditional logits} & \\ 
	Caption Embedding & $nef$\\
	Concat w/ replicated caption embedding & $8*ndf+nef \times 4 \times 4$ \\
	Block3x3\_leakyReLU & $8*ndf \times 4 \times 4$\\
	Conv(k=4, s=4, p=0, b=True) & $1$\\
	Sigmoid & $1$ \\
	\\
	\textbf{Discriminator $64 \times 64$} & \\
	Input tensor & $3 \times 64 \times 64$ \\
	Down Block & $ndf \times 32 \times 32$ \\
	Down Block & $2*ndf \times 16 \times 16$ \\
	Down Block & $4*ndf \times 8 \times 8$\\
	Down Block & $8*ndf \times 4 \times 4$\\ 
	\textit{Unconditional logits} & \\  
	Conv(k=4, s=4, p=0, b=True) & $1$ \\
	Sigmoid & $1$ \\ 
	\textit{Conditional logits} & \\ 
	Caption Embedding & $nef$\\
	Concat w/ replicated caption embedding & $8*ndf+nef \times 4 \times 4$ \\ 
	Block3x3\_leakyReLU & $8*ndf \times 4 \times 4$\\
	Conv(k=4, s=4, p=0, b=True) & $1$\\
	Sigmoid & $1$ \\
	\bottomrule
\end{tabular}%
}
\par\end{center}%
\end{minipage}
\par\end{center}%
\end{minipage}
\par\end{centering}
\end{table}

\begin{table}[t]
\caption{Network details of our countermodel. Some components which are not mentioned here such as text encoder, image encoder, DAMSM, its settings can be found in AttnGAN~\cite{xu2018attngan}. In the tables, k = kernel size, s = stride, p = padding, b = bias.}
\label{tab:counter_model_settings}
\begin{centering}
\noindent\begin{minipage}[t]{1\columnwidth}%
\begin{center}
\begin{minipage}[t]{0.45\columnwidth}%
\begin{center}
\label{sub_tab:counter_model_generator}
(a) Generator 
\resizebox{\columnwidth}{!}{%
\begin{tabular}{l | c} 
	\toprule
	Module & Output shape / Details \\
	\midrule
	\textbf{Up Block} & see Table~\ref{sub_tab:attn_gan_plus_plus_generator}\\
	\\
	\textbf{Residual Block} & see Table~\ref{sub_tab:attn_gan_plus_plus_generator} \\
	\\
	\textbf{Spatial Attention Layer} & see AttnGAN \\
	\\
	\textbf{Conditional Augumentation} & see AttnGAN \\
	\\
	\textbf{Generator $4 \times 4$} & \\
	\textit{Input} & \\
	Input noise & $nzf$\\
	Caption Embedding  & $nef$ \\
	\textit{Computation} & \\
	Conditional Augumentation on caption embedding & $ncf$ \\
	Concat w/ noise & $ncf + nef$ \\ 
	Linear(b=False) &  $ngf * 16 * 4 * 4 * 2$\\
	BatchNorm1D & No change shape \\
	Gated Linear Unit (GLU) & $ngf * 16 * 4 * 4 * 1$ \\
	Reshape & $16 * ngf \times 4 \times 4$ \\ 
	Conv(k=3, s=1, p=1, b=False) & $3 \times 4 \times 4$ \\
	Tanh & No change shape \\
	\\
	\textbf{Generator $8 \times 8$} & \\
	Up Block  & $8 * ngf \times 8 \times 8$ \\
	Conv(k=3, s=1, p=1, b=False) & $3 \times 8 \times 8$ \\
	Tanh & No change shape \\
	\\
	\textbf{Generator $16 \times 16$} & \\
	Up Block  & $4 * ngf \times 16 \times 16$ \\
	Conv(k=3, s=1, p=1, b=False) & $3 \times 16 \times 16$ \\
	Tanh & No change shape \\
	\\
	\textbf{Generator $32 \times 32$} & \\
	Up Block  & $2 * ngf \times 32 \times 32$ \\
	Conv(k=3, s=1, p=1, b=False) & $3 \times 32 \times 32$ \\
	Tanh & No change shape \\
	\\
	\textbf{Generator $64 \times 64$} & \\
	Up Block  & $ngf \times 64 \times 64$ \\
	Conv(k=3, s=1, p=1, b=False) & $3 \times 64 \times 64$ \\
	Tanh & No change shape \\
	\\
	\textbf{Generator $128 \times 128$} & \\
	\textit{Input} & \\
	Previous hidden features & $ngf \times 64 \times 64 $ \\
	Word Mask & $word\_num$ \\
	Word features & $nef \times word\_num$ \\ 
	\textit{Computation} & \\
	Spatial Attention Layer & $ngf \times 64 \times 64 $ \\
	Residual Block $\times$ residual\_num & $ngf \times 64 \times 64$ \\
	Concat w/ previous hidden features  & $2 * ngf \times 64 \times 64$ \\
	Up Block $ngf \times 128 \times 128$ \\
	Conv(k=3, s=1, p=1, b=False) & $3 \times 128 \times 128$ \\
	Tanh & No change shape \\
	\\
	\textbf{Generator $256 \times 256$} & \\
	\textit{Input} & \\
	Previous hidden features & $ngf \times 128 \times 128 $ \\
	Word Mask & $word\_num$ \\
	Word features & $nef \times word\_num$ \\ 
	\textit{Computation} & \\
	Spatial Attention Layer & $ngf \times 128 \times 128 $ \\
	Residual Block $\times$ residual\_num & $ngf \times 128 \times 128$ \\
	Concat w/ previous hidden features  & $2 * ngf \times 128 \times 128$ \\
	Up Block  & $ngf \times 256 \times 256$ \\
	Conv(k=3, s=1, p=1, b=False) & $3 \times 256 \times 256$ \\
	Tanh & No change shape \\
	\bottomrule
\end{tabular}%
}
\par\end{center}%
\end{minipage}\quad{}%
\hfill
\begin{minipage}[t]{0.45\columnwidth}%
\begin{center}
\label{sub_tab:counter_model_discriminator}
(b) Discriminator 
\resizebox{\columnwidth}{!}{%
\begin{tabular}{l | c} 
	\toprule
	Module & Output shape / Details \\
	\midrule
	\textbf{Block3x3\_leakyReLU} & see Table~\ref{sub_tab:attn_gan_plus_plus_discriminator}\\
	\\
	\textbf{DisGeneralConvBlock} & \\
    \emph{Params: in\_planes, concat\_planes, out\_planes} \\
    MinibatchStdDev (see~\cite{karras2017progressive,karras2019style}) & in\_planes + concat\_planes $\times$ h $\times$ w  \\
    Block3x3\_leakyRelu & in\_planes $\times$ h $\times$ w\\
    Block3x3\_leakyRelu & out\_planes $\times$ h $\times$ w \\
    AvgPool2d(k=2) & out\_planes $\times$ h/2 $\times$ w/2\\
    \\
	\textbf{Discriminator} & \\
	\textit{Input} & \\
	Caption Embedding & $nef$\\
	Image scale $4 \times 4$ & $3 \times 4 \times 4$ \\
	Image scale $8 \times 8$ & $3 \times 8 \times 8$ \\
	Image scale $16 \times 16$ & $3 \times 16 \times 16$ \\
	Image scale $32 \times 32$ & $3 \times 32 \times 32$ \\
	Image scale $64 \times 64$ & $3 \times 64 \times 64$ \\
	Image scale $128 \times 128$ & $3 \times 128 \times 128$ \\
	Image scale $256 \times 256$ & $3 \times 256 \times 256$ \\
	\textit{Computation} & \\ 
	Image scale $256 \times 256$ & $3 \times 256 \times 256$ \\
	Conv(k=1, s=1, p=0, b=True) & $ndf \times 256 \times 256$ \\
	DisGeneralConvBlock($ndf$, $1$, $2*ndf$) & $ndf * 2 \times 128 \times 128$ \\
	Concat w/ Image scale $128 \times 128$ & $ndf * 2 + 3 \times 128 \times 128$ \\
	DisGeneralConvBlock($2*ndf$, $4$, $4*ndf$) & $4*ndf \times 64 \times 64$ \\
	Concat w/ Image scale $64 \times 64$ & $4*ndf + 3 \times 64 \times 64$ \\
	DisGeneralConvBlock($4*ndf$, $4$, $8*ndf$) & $8*ndf \times 32 \times 32$\\
	Concat w/ Image scale $32 \times 32$ & $8*ndf + 3 \times 32 \times 32$ \\
	DisGeneralConvBlock($8*ndf$, $4$, $8*ndf$) & $8*ndf \times 16 \times 16$\\
	Concat w/ Image scale $16 \times 16$ & $8*ndf + 3 \times 16 \times 16$ \\
	DisGeneralConvBlock($8*ndf$, $4$, $8*ndf$) & $8*ndf \times 8 \times 8$\\
	Concat w/ Image scale $8 \times 8$ & $8*ndf + 3 \times 8 \times 8$ \\
	DisGeneralConvBlock($8*ndf$, $4$, $8*ndf$) & $8*ndf \times 4 \times 4$ \\
	\textit{Unconditional logits} & \\  
	Conv(k=4, s=4, p=0, b=True) & $1$ \\
	Sigmoid & $1$ \\ 
	\textit{Conditional logits} & \\ 
	Caption Embedding & $nef$\\
	Concat w/ replicated caption embedding & $8*ndf+nef \times 4 \times 4$ \\
	Block3x3\_leakyReLU & $8*ndf \times 4 \times 4$\\
	Conv(k=4, s=4, p=0, b=True) & $1$\\
	Sigmoid & $1$ \\
	\bottomrule
\end{tabular}%
}
\par\end{center}%
\end{minipage}
\par\end{center}%
\end{minipage}
\par\end{centering}
\end{table}

\begin{table}[t]
\centering
\caption{Training settings of both AttnGAN++ and counter model. Most of settings in evaluation process is the same with training process except word\_num. In the evaluation process, word\_num=$25$ for the CUB~\cite{welinder2010caltech} dataset and word\_num=$20$ for the MS-COCO~\cite{lin2014microsoft} dataset.}
\label{tab:training_settings}
\resizebox{\linewidth}{!}{%
\begin{tabular}{c | c | c} 
\toprule
Dataset & CUB~\cite{welinder2010caltech} & MS-COCO~\cite{lin2014microsoft} \\
\midrule
Optimizer & Adam($\beta_{1}=0.5$, $\beta_{2}=0.999$) & Adam($\beta_{1}=0.5$, $\beta_{2}=0.999$)\\
Generator (G) Learning Rate & $0.0002$ & $0.0002$ \\
Discriminator (D) Learning Rate & $0.0002$ & $0.0002$ \\
G/D Update & $1:1$ & $1:1$ \\
$\gamma_{1}$ & 4.0 & 4.0 \\
$\gamma_{2}$ & 5.0 & 5.0 \\
$\gamma_{3}$ & 10.0 & 10.0\\
$\lambda$ & 5.0 & 50.0\\
residual\_num & 2 & 3 \\
ngf & $64$ & $64$ \\
ndf & $32$ & $32$ \\
nef & $256$  & $256$\\
nzf & $100$  & $100$\\
ncf & $100$  & $100$\\
max\_epochs & $800$ & $200$ \\
word\_num & $18$ & $12$ \\
\bottomrule
\end{tabular}
}
\end{table}


\end{document}